\PassOptionsToPackage{disable}{endfloat}
\documentclass[pdflatex,sn-mathphys-num]{sn-jnl}% Math and Physical Sciences Numbered Reference Style
\usepackage{array}
\usepackage{graphicx}%
\usepackage{placeins}%
\usepackage{multirow}%
\usepackage{amsmath,amssymb,amsfonts}%
\usepackage{amsthm}%
\usepackage{mathrsfs}%
\usepackage[title]{appendix}%
\usepackage{xcolor}%
\usepackage{textcomp}%
\usepackage{manyfoot}%
\usepackage{booktabs}%
\usepackage{algorithm}%
\usepackage{algorithmicx}%
\usepackage{algpseudocode}%
\usepackage{listings}%
\theoremstyle{thmstyleone}%
\theoremstyle{thmstyletwo}%

\theoremstyle{thmstylethree}%

\begin{document}
\makeatletter
\@ifundefined{efloat@restorefloats}{}{\efloat@restorefloats}
\makeatother

\title[Article Title]{\textbf{PASS}: A Trustworthy Precision Approach Simulation System for CAT III Integrity Algorithm Development}


\author[1]{\fnm{Lang} \sur{Zhu}}\author*[1]{\fnm{Xin} \sur{Zhang}}\email{xin.zhang@sjtu.edu.cn}\author[1]{\fnm{Yulong} \sur{Sun}}\author[2]{\fnm{Nan} \sur{Liu}}\author[2]{\fnm{Jiahe} \sur{Shen}}\author[2]{\fnm{Xiaodong} \sur{Zhang}}\author[1]{\fnm{Xingqun} \sur{Zhan}}\affil*[1]{\orgname{Shanghai Jiao Tong University}, \orgaddress{\city{Shanghai}, \country{China}}}\affil[2]{\orgname{Flight Automatic Control Research Institute}, \orgaddress{\city{Xi'an}, \country{China}}}


\abstract{For fixed-wing aircraft, reliable localization during approach and landing is critical to flight safety. Visual--inertial navigation is promising, but aircraft navigation also requires integrity: trustworthy outputs and timely warnings when they should not be used. Real approach data are costly to collect, while many visual errors and failures required for integrity algorithm development are difficult to reproduce safely. We present \textbf{PASS}, a trustworthy Precision Approach Simulation System for CAT III integrity algorithm development. In a unified temporal and geometric framework, \textbf{PASS} integrates configurable airport environments, prescribed trajectories, inertial simulation, calibrated stereo images, and continuous runway-relative ground truth. It supports navigation and integrity research through controllable sensor errors and environmental variations, computable visual projection errors, deterministic runway geometry priors, and condition-labeled data for statistical error modeling. We further propose Sim-to-Real Scene Trustworthiness (\textbf{SRST}), a measure based on DINOv2 feature distributions and UMAP, to quantify sim-to-real closeness. Experiments with representative stereo--inertial estimators demonstrate the usability of the generated sequences and quantify localization errors across algorithms and environments. \textbf{PASS} provides a controllable foundation for developing and evaluating visual--inertial navigation and integrity algorithms for fixed-wing precision approaches up to CAT III requirements.}

\keywords{precision approach simulation system; Visual--Inertial navigation integrity; Simulation Trustworthiness;}



\maketitle

\section{Introduction}\label{sec1}

Reliable runway-relative localization is essential for autonomous fixed-wing approach and landing, particularly when conventional landing aids or Global Navigation Satellite System (GNSS) signals are unavailable, degraded, or unreliable. Cameras and inertial sensors provide complementary geometric and motion information, making visual--inertial navigation a promising solution for runway perception, aircraft pose estimation, and approach navigation \cite{Marianandam2014RunwayAlignment,Tsapparellas2024RunwayLocalization,Ferreira2025LandingGuidance}. In safety-critical aircraft navigation, however, positioning accuracy alone is insufficient. Navigation integrity is also required to determine whether an estimated state can be trusted and to identify misleading information before it compromises the operation. As visual measurements become increasingly involved in aircraft localization, integrity monitoring for vision-aided navigation is therefore receiving growing attention \cite{Sun2026IntegrityVINS}.

Developing visual--inertial navigation and integrity algorithms requires more than synchronized sensor measurements. Navigation algorithms depend on continuous imagery, inertial measurements, calibration parameters, and reliable trajectory references, whereas integrity algorithms additionally require controllable error excitation, truth-referenced visual errors (differences between detected runway features and their projections from independent pose and runway geometry truth), deterministic geometric priors, and condition-labeled observations for uncertainty modeling and error bounding. These reference errors are needed to characterize and bound visual-measurement uncertainty for downstream computation of navigation integrity risk; \textbf{PASS} provides the requisite data rather than calculating that risk itself. Acquiring this information from real fixed-wing approaches is costly and constrained by operational and safety requirements. In particular, adverse visual conditions, sensor disturbances, prior information inconsistencies, and rare failure events cannot be introduced freely or reproduced systematically during flight tests. Existing runway-oriented datasets such as LARD and BARS improve access to labeled approach imagery, but primarily provide individual images and two-dimensional annotations rather than continuous calibrated visual--inertial sequences \cite{Ducoffe2023LARD,Chen2023BARS}. They also lack the controlled disturbances, independent error references, runway priors, and condition labels needed to model measurement covariance, assess residual consistency, construct conservative error bounds, and determine the operating conditions under which a visual error model remains trustworthy. Existing resources are therefore insufficient for reproducible research that connects runway perception with visual--inertial navigation and integrity evaluation throughout fixed-wing precision approaches.

Simulation provides a practical means of generating repeatable sensor observations and exact geometric references under controllable airport, environmental, and sensor conditions. Nevertheless, two related questions must be addressed before simulated data can support downstream integrity analysis. First, fully simulated system must expose the information needed to construct and evaluate integrity models rather than merely reproduce nominal visual--inertial measurements. Second, the generated observations must be sufficiently trustworthy with respect to real approach imagery, because visually plausible rendering alone does not establish the applicability of an error model developed in simulation. The rendered stereo images, inertial measurements, calibration parameters, coordinate transformations, runway geometry, and trajectory ground truth must therefore remain mutually consistent, while the remaining sim-to-real discrepancy must be quantified explicitly.

To address these requirements, we present \textbf{PASS}: A Trustworthy Precision Approach Simulation System designed to support CAT III integrity algorithm development. \textbf{PASS} couples prescribed approach trajectories and inertial simulation based on the Precise Strapdown Inertial Navigation System (PSINS) toolbox~\cite{Yan2024PSINS} with configurable multi-airport environments and calibrated stereo imaging in Unreal Engine. A shared time base and an explicit transformation chain connect the global, runway, aircraft, inertial, camera, and image frames, allowing the system to generate continuous visual--inertial observations together with traceable runway-relative ground truth. \textbf{PASS} is not itself an integrity monitoring or navigation algorithm and does not directly output a protection level. Instead, it provides a controlled simulation and data interface through which navigation algorithms can be exercised and the information required by downstream integrity models can be generated, measured, and statistically characterized.

The main contributions of this work are summarized as follows:

\begin{itemize}
    \item We develop \textbf{PASS}: A Trustworthy Precision Approach Simulation System that integrates PSINS-based trajectory and inertial simulation, configurable airport environments, and calibrated stereo-image generation. Its modular architecture, shared temporal basis, traceable sensor configuration, and explicit coordinate transformation chain maintain consistency across the generated multimodal data while allowing airport, trajectory, environment, and sensor parameters to be configured independently.

    \item We enable \textbf{PASS} to provide structured support information for the development and evaluation of visual--inertial navigation integrity algorithms. This information comprises controllable sensor error excitation and environmental variation, computable visual projection errors referenced to ground truth, deterministic runway geometry priors, and condition-labeled data for statistical error modeling. Together, these capabilities support visual error distribution modeling, dynamic covariance estimation, residual construction, conservative error bounding, and integrity analysis across operating conditions while keeping the downstream integrity algorithm independent of the simulation framework.

    \item We introduce \emph{Sim-to-Real Scene Trustworthiness} (SRST), a stage-conditioned measure for small samples of real and simulated images acquired during airport approach. The measure is designed to mirror a human comparative scene judgment: a simulated image is assessed against both the corresponding category in the real-image set and competing categories. Using frozen DINOv2 features and a joint two-dimensional UMAP representation constructed separately for each approach stage, SRST quantifies category-relative scene correspondence across coastal, mountainous, desert, and urban airports without requiring image-level pairs.
\end{itemize}

The remainder of this paper is organized as follows. Section~\ref{sec2} reviews datasets and simulation resources relevant to fixed-wing runway perception, visual--inertial navigation, and downstream integrity algorithm development. Section~\ref{sec3} presents the \textbf{PASS} architecture, its modular simulation subsystems, and the integrity support information provided by the framework. Section~\ref{sec4} evaluates the functional usability of the generated sequences using representative stereo--inertial estimators. Section~\ref{sec5} concludes the paper.

The \textbf{PASS} dataset will be made publicly available upon publication at \url{https://github.com/zhulang195/PASS_dataset}.

\section{Related Work}\label{sec2}
The value of an approach dataset for navigation integrity depends both on the information it provides and on the credibility of its simulated observations. Synchronized stereo images and inertial measurements must be accompanied by trustworthy sensor parameters, runway-relative trajectory references, and explicit runway geometry to connect perception with navigation-state estimation. Developing downstream integrity algorithms additionally requires controllable error sources, independent truth references, deterministic geometric priors, and condition-labeled observations from which measurement error distributions, dynamic covariance models, residual tests, and conservative error bounds can be constructed \cite{Sun2026IntegrityVINS}. Sim-to-real assessment addresses a complementary question: whether the visual organization of simulated approach scenes corresponds to that of real imagery under the available reference information. This section therefore reviews both runway and navigation datasets and methods for assessing simulation credibility.

Runway-oriented image datasets are closely related to the visual conditions encountered during fixed-wing approach. LARD provides predominantly synthetic front-view runway images with automatically projected runway-corner labels and supplements them with manually annotated real images \cite{Ducoffe2023LARD}. LARD~2.0 extends this concept through multiple virtual globes and flight simulators, incorporates complex multi-runway airports, and introduces an operational-design-domain-aware detection benchmark \cite{Bougacha2026LARD2}. BARS and RLD similarly provide runway imagery and instance-level segmentation annotations under different terrain, weather, and viewing conditions \cite{Chen2023BARS,Wang2024VALNet}. These datasets have advanced runway detection, segmentation, and keypoint estimation, particularly under small-target, perspective-change, and adverse-appearance conditions. Their published data units, however, are primarily individual images and two-dimensional annotations. They do not provide continuous calibrated stereo--inertial measurements, runway-relative aircraft states, or time-aligned error references. Consequently, they can support runway perception but cannot independently support continuous visual--inertial navigation or integrity analysis throughout an approach.

Synthetic datasets and simulation platforms provide an alternative means of generating repeatable observations under controllable conditions. UnrealCV enables programmatic camera control and ground-truth extraction in Unreal Engine, while AirSim provides configurable vehicles, sensors, and hardware-in-the-loop interfaces \cite{Qiu2016UnrealCV,Shah2017AirSim}. Mid-Air releases simulated camera, inertial, positioning, depth, and semantic data under multiple climate conditions, whereas TartanAir provides stereo imagery, camera poses, depth, segmentation, and optical flow across diverse environments and motion patterns \cite{Fonder2019MidAir,Wang2020TartanAir}. These resources demonstrate the scalability of synthetic-data generation. Nevertheless, their scenes, trajectories, coordinate frames, and reference quantities are designed primarily for general visual navigation or low-altitude multirotor operation rather than integrity evaluation, the standard compulsory procedure for precision approach. Although some platforms are configurable at the simulator level, their released data and interfaces do not natively organize runway geometry, runway-relative states, calibrated stereo--inertial sensing, controllable prior information inconsistencies, and in particular, truth-referenced visual errors within a single approach-oriented framework. Substantial redevelopment would therefore be required to use them for systematic visual error modeling and navigation integrity studies. Moreover, visual plausibility alone cannot establish simulation credibility; the internal consistency of the sensor models and the remaining discrepancy between simulated and real approach imagery must also be evaluated explicitly.

Having the dataset for integrity evaluation, a natural question is the trustworthiness of the simulator. Quantitative assessment of visual sim-to-real correspondence is commonly formulated as a comparison between real and synthetic image distributions in a fixed feature space. The Fréchet Inception Distance (FID) approximates inception features by Gaussian distributions and compares their means and covariances \cite{Heusel2017FID}, whereas the Kernel Inception Distance (KID) applies a polynomial-kernel maximum mean discrepancy to the same feature family \cite{Binkowski2018KID}. More generally, maximum mean discrepancy provides a kernel-based two-sample statistic without requiring an explicit parametric distribution model \cite{Gretton2012MMD}. Recent work has replaced inception features with representations from more broadly pretrained visual encoders; for example, CMMD combines CLIP features with a Gaussian-kernel discrepancy \cite{Jayasumana2024CMMD}, while sim-to-real studies have also used distances between fixed pretrained features to characterize appearance gaps and their relationship with downstream performance \cite{Vanherle2024Sim2RealAugmentations}. These approaches make distribution-level comparison possible for unpaired images. Their values, however, depend on the feature extractor, preprocessing, sample composition, and discrepancy definition. FID additionally exhibits finite-sample bias \cite{Chong2020UnbiasedFID}, and a low aggregate discrepancy does not by itself show that the synthetic data preserve individual scene categories or cover the relevant variation in the real domain.

Distributional proximity should therefore be distinguished from conditional correspondence, spatial correspondence, and task-level predictivity. Feature-space precision and recall separate the extent to which generated samples remain within the support of real data from the extent to which they cover real-data variation \cite{Kynkaanniemi2019PrecisionRecall}. Conditional evaluation further addresses cases in which images may appear realistic overall while remaining inconsistent with their intended conditions \cite{Soloveitchik2021CFID}. Stronger forms of correspondence require additional reference information. Perceptual distances such as Learned Perceptual Image Patch Similarity (LPIPS) assume images with sufficiently comparable content or viewpoint \cite{Zhang2018LPIPS}, whereas visual place-recognition measures such as Recall@K require reliable place identities and a cross-domain retrieval gallery \cite{Keetha2024AnyLoc}. Downstream task accuracy and sim-to-real predictivity provide more direct evidence of application utility, but require a defined perception or navigation system that can be evaluated in both domains \cite{Kadian2020Sim2RealPredictivity}. These evaluation modes consequently answer different questions and cannot be interpreted as interchangeable measures of simulation credibility.

The remaining research gap is therefore twofold. First, existing runway-image datasets and general-purpose synthetic navigation resources do not jointly provide the calibrated visual--inertial observations, continuous runway-relative ground truth, deterministic geometric priors, controllable disturbances, and condition-labeled error references required for integrity  algorithm development. Second, the inclusion of visually plausible synthetic imagery is rarely accompanied by an explicit evaluation of how its scene organization corresponds to real fixed-wing approach imagery under the available category and approach-stage information. Reporting synthetic and real examples, or demonstrating simulator configurability, does not by itself resolve either requirement.

\textbf{PASS} addresses these complementary gaps within a common precision approach framework. It integrates calibrated stereo--inertial observations, runway-relative state truth, configurable operating conditions, integrity support information, and an explicit stage-conditioned assessment of sim-to-real scene trustworthiness. Table~\ref{tab:dataset_comparison} summarizes these distinctions. \textbf{PASS} does not implement a particular navigation or integrity monitoring algorithm, and SRST is not presented as a universal measure of visual realism or a certified safety threshold. Instead, the system provides controlled observations and traceable reference information through which navigation and integrity algorithms can be developed and evaluated, while SRST supplies bounded evidence that the simulated visual observations preserve the intended category-level organization of the evaluated real approach scenes.

\begin{table*}[t]
\caption{Comparison of representative runway-oriented datasets and synthetic simulation resources for fixed-wing precision approach navigation and integrity research.}
\label{tab:dataset_comparison}
\centering
\scriptsize
\resizebox{\textwidth}{!}{%
\begin{tabular}{llcccccc}
\hline
Resource &
Type &
\shortstack{Fixed-wing\\runway approach} &
\shortstack{Continuous\\stereo--IMU} &
\shortstack{Runway-relative\\ground truth} &
\shortstack{Configurable errors\\or conditions} &
\shortstack{Integrity support\\information} &
\shortstack{Sim-to-real\\credibility assessment} \\
\hline

LARD v1 \cite{Ducoffe2023LARD}
& Dataset
& Yes
& No
& No
& No
& No
& Partial \\

LARD v2 \cite{Bougacha2026LARD2}
& Dataset
& Yes
& No
& No
& No
& No
& Partial \\

BARS \cite{Chen2023BARS}
& Dataset
& Yes
& No
& No
& No
& No
& No \\

RLD/VALNet \cite{Wang2024VALNet}
& Dataset
& Yes
& No
& No
& No
& No
& No \\

UnrealCV \cite{Qiu2016UnrealCV}
& Simulation framework
& No
& No
& No
& Partial
& No
& No \\

AirSim \cite{Shah2017AirSim}
& Simulation framework
& No
& Supported
& No
& Supported
& No
& No \\

Mid-Air \cite{Fonder2019MidAir}
& Synthetic dataset
& No
& Yes
& No
& Preset
& No
& No \\

TartanAir \cite{Wang2020TartanAir}
& Synthetic dataset
& No
& No
& No
& Preset
& No
& No \\

\textbf{PASS} (ours)
& Simulation system
& Yes
& Yes
& Yes
& Yes
& Yes
& Yes \\
\hline
\end{tabular}%
}

\vspace{3pt}
\parbox{0.98\textwidth}{\footnotesize
``Supported'' indicates that the corresponding capability can be implemented
through the simulation platform but is not provided as an integrated,
runway-oriented data product. ``Preset'' denotes condition diversity contained
in released sequences rather than user-controlled error excitation.
Integrity support information refers to the combined availability of
controllable sensor or environmental disturbances, truth-referenced visual
errors, deterministic runway geometry priors, and condition-labeled data for
statistical error modeling. ``Partial'' sim-to-real assessment denotes the
inclusion or comparison of synthetic and real imagery without an explicit
quantitative credibility indicator conditioned on approach stage.
}
\end{table*}

\section{\textbf{PASS}: A Trustworthy Precision Approach Simulation System for CAT III Integrity Algorithm Development}\label{sec3}

We propose \textbf{PASS}: A Trustworthy Precision Approach Simulation System designed to support visual--inertial navigation and CAT III integrity algorithm development. This section first presents the overall architecture and modular components of \textbf{PASS}. It then describes the controlled error conditions, independent truth references, geometric priors, and condition-labeled observations provided for downstream integrity algorithm development. Finally, it introduces a credibility-oriented sim-to-real quality assessment based on image distributions conditioned on approach stage in a high-dimensional feature space.

\subsection{System Architecture}\label{sec3_1}

\begin{figure*}[t]
    \centering
    \includegraphics[width=0.96\textwidth]{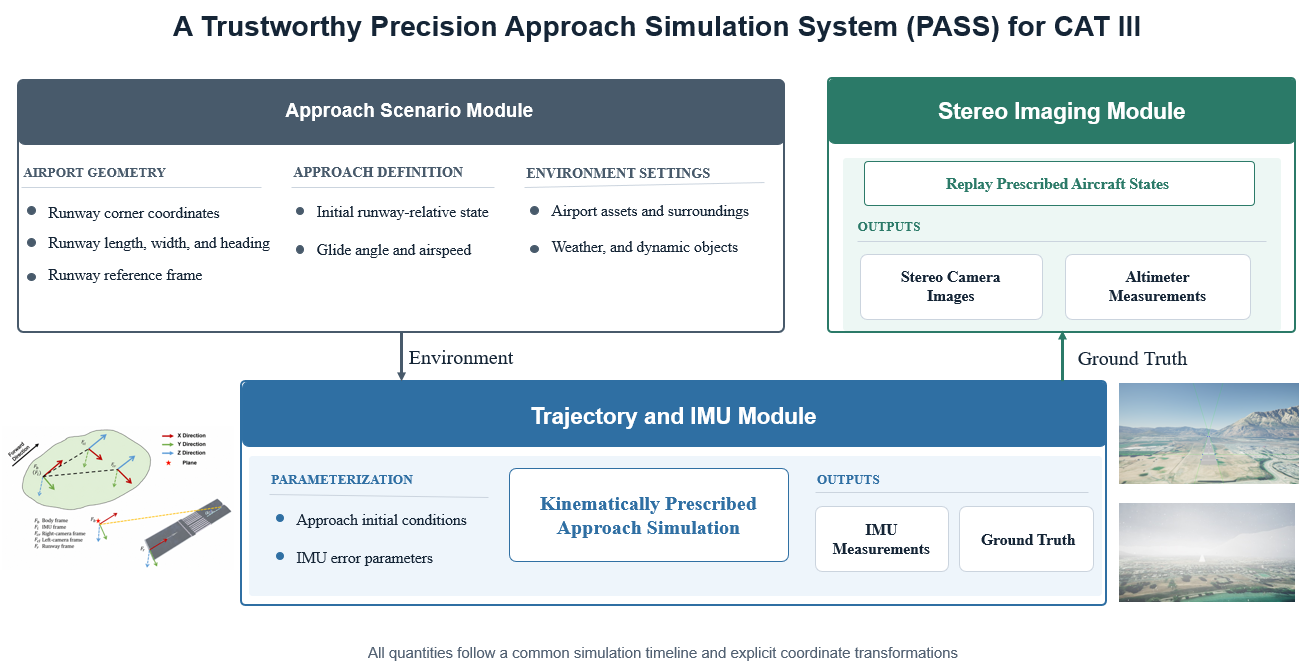}
    \caption{Overall architecture and information flow of \textbf{PASS}. The Approach Scenario Module defines the runway geometry, operating conditions, and controllable disturbances required for downstream integrity analysis. The Trajectory and IMU Module generates the prescribed aircraft states, runway-relative ground truth, and simulated inertial measurements. The Stereo Imaging Module replays the prescribed motion and records synchronized stereo images. A shared time base and coordinate transformation chain maintain temporal and geometric consistency across the three modules.}
    \label{fig:framework_overview}
\end{figure*}

\textbf{PASS} comprises three tightly coupled modules. The Approach Scenario Module constructs configurable multi-airport and multi-runway environments and defines the operating conditions and sensor configurations. The Precise Strapdown Inertial Navigation System (PSINS)-based Trajectory and IMU Module~\cite{Yan2024PSINS} generates the prescribed aircraft motion, navigation states, and inertial measurements. The Stereo Imaging Module then replays the prescribed states in Unreal Engine and records synchronized stereo images using calibrated virtual cameras. As illustrated in Fig.~\ref{fig:framework_overview}, a shared time base and an explicit coordinate transformation chain maintain temporal and geometric consistency throughout this process.

The resulting sequences integrate three complementary types of information: integrity support conditions and runway geometry priors, continuous trajectory and runway-relative ground truth accompanied by inertial measurements, and synchronized visual observations. These data are organized with the associated calibration and coordinate frame parameters in a common sequence structure, providing a consistent basis for visual--inertial navigation experiments and measurement error characterization for downstream integrity analysis.

The interfaces between the modules are listed in Table~\ref{tab:framework_interfaces}. The numerical trajectory files serve two purposes. They are released as ground truth, and they drive the virtual platform during image generation. This shared source prevents the visual trajectory from being generated independently of the IMU and pose records. 

\begin{table*}[t]
\caption{Inputs and outputs of the three modules in \textbf{PASS} shown in Fig.~\ref{fig:framework_overview}.}
\label{tab:framework_interfaces}
\centering
\footnotesize % 缩小字体
\setlength{\tabcolsep}{4pt} % 缩小列间距

% 在首尾加上 @{}，消除表格默认的左右外边距，使得所有内容（包括圆点）严格左对齐
\begin{tabular}{@{}p{0.28\textwidth} >{\raggedright\arraybackslash}p{0.30\textwidth} >{\raggedright\arraybackslash}p{0.30\textwidth}@{}}
\hline
\multicolumn{1}{c}{Module} & \multicolumn{1}{c}{Main inputs} & \multicolumn{1}{c}{Main outputs} \\ 
\hline 
\noalign{\vskip 1.5ex} % 标题与内容之间保留距离
Approach Scenario Module & 
\begin{tabular}[t]{@{}l@{}}
\raisebox{0.2ex}{\tiny\textbullet}~Airport Geometry \\
\raisebox{0.2ex}{\tiny\textbullet}~Approach Definition \\
\raisebox{0.2ex}{\tiny\textbullet}~Environment Settings

\end{tabular} & 
\begin{tabular}[t]{@{}l@{}}
\raisebox{0.2ex}{\tiny\textbullet}~Approach conditions \\
\raisebox{0.2ex}{\tiny\textbullet}~Environment config
\end{tabular} \\
Trajectory and IMU Module & 
\begin{tabular}[t]{@{}l@{}}
\raisebox{0.2ex}{\tiny\textbullet}~Upstream geometry \\
\raisebox{0.2ex}{\tiny\textbullet}~IMU error parameters
\end{tabular} & 
\begin{tabular}[t]{@{}l@{}}
\raisebox{0.2ex}{\tiny\textbullet}~Continuous trajectory \\
\raisebox{0.2ex}{\tiny\textbullet}~Simulated IMU data
\end{tabular} \\
Stereo Imaging Module & 
\begin{tabular}[t]{@{}l@{}}
\raisebox{0.2ex}{\tiny\textbullet}~Upstream states \\
\raisebox{0.2ex}{\tiny\textbullet}~Camera calibration \\
\end{tabular} & 
\begin{tabular}[t]{@{}l@{}}
% 这里不再使用 ~~~ 缩进，而是给两行都加上圆点，保证圆点严格左对齐
\raisebox{0.2ex}{\tiny\textbullet}~Altimeter Measurements \\
\raisebox{0.2ex}{\tiny\textbullet}~Stereo Camera Images

\end{tabular} \\
\hline
\end{tabular}
\end{table*}

\subsection{Modular Subsystems}\label{sec3_2}

The three-module architecture is implemented through three coordinated subsystems. This separation allows the visual environment, approach trajectory, sensor configuration, and geometric references to be adjusted independently while preserving a unified data-generation interface and cross-modal consistency.

\begin{figure}[t]
    \centering
    \includegraphics[width=0.68\linewidth]{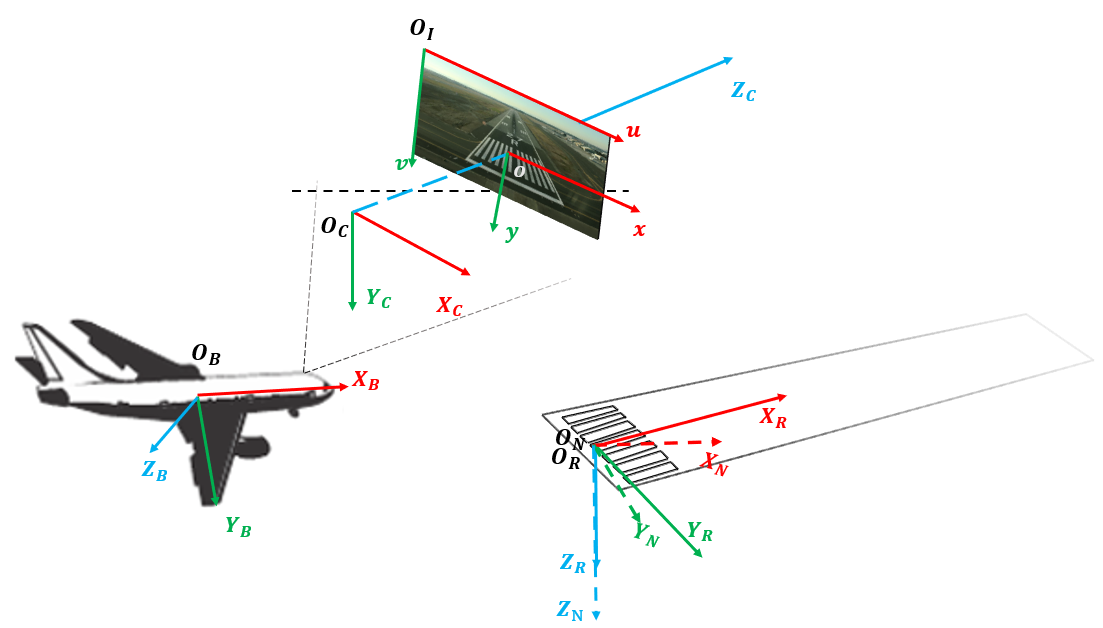}
    \caption{Coordinate framework of \textbf{PASS}. The transformation chain connects the geodetic and local navigation frames to the runway, aircraft body/IMU, stereo camera, and image frames, providing a common geometric basis for multimodal data generation.}
    \label{fig:coordinate_frames}
\end{figure}

A shared coordinate framework connects the trajectory, inertial, visual, and runway-referenced information generated by \textbf{PASS}, as summarized in Fig.~\ref{fig:coordinate_frames}.

\subsubsection{Approach Scenario Module}\label{sec3_2_1}

\paragraph{Airport and runway construction.}

We developed a parameterized, runway-centered visual environment in Unreal Engine~5.1.1 to reproduce fixed-wing precision approach scenes under controlled geometric conditions. Each airport is represented by an airport-specific set of runway parameters, including the runway location, dimensions, orientation, surface appearance, markings, and associated visual aids. These elements were modeled with reference to the physical-characteristic and visual-aid provisions of ICAO Annex~14, Volume~I~\cite{ICAO2022Annex14}. Rather than treating the runway as part of the background imagery, \textbf{PASS} represents its principal geometric elements explicitly, allowing the rendered runway to remain consistent with the runway reference frame and the geometric priors used by downstream algorithms.

\paragraph{Controlled visual conditions.}\label{sec3_2_2}

We further parameterized the principal environmental factors that affect runway visibility and visual-feature extraction. Illumination is varied through the solar and atmospheric states provided by CesiumSunSky and Ultra Dynamic Sky, including representative morning, noon, and dusk conditions. These factors can be varied individually or in combination while the prescribed trajectory, camera calibration, and runway geometry remain unchanged. This controlled separation between visual conditions and geometric configuration enables differences in the resulting observations to be attributed to defined environmental changes rather than to unintended variations in aircraft motion or sensor geometry.

\subsubsection{Trajectory and IMU Module}\label{sec3_2_3}

To establish a controlled reference motion for precision approach, we parameterize each trajectory using the geodetic initial and terminal positions together with the prescribed fixed-wing approach state, including the initial attitude, velocity, and glide-path conditions. The PSINS-based simulation propagates these boundary and motion conditions to generate a continuous position, velocity, and attitude trajectory, together with the corresponding ideal angular and velocity increments. The same state history is used as the reference trajectory for inertial-data generation and as the motion input for visual replay in Unreal Engine. This common motion basis ensures that the simulated inertial measurements, rendered aircraft motion, and runway-relative pose ground truth describe the same prescribed approach.

The ideal inertial increments are subsequently converted into simulated IMU measurements through a configurable error model. The model incorporates both deterministic and stochastic error components, including constant gyroscope and accelerometer biases, random walk, first-order correlated bias, scale-factor error, and inter-axis misalignment. The sampling interval and error parameters are specified independently of the reference trajectory, allowing different IMU characteristics to be evaluated under an identical approach motion. The principal configurable quantities are summarized in Table~\ref{tab:imu_error}. The module therefore supplies both continuous motion truth and controlled inertial-error realizations for subsequent visual--inertial navigation and downstream integrity analyses.

\begin{table*}[t]
\caption{Parameterization of the PSINS-based IMU simulation model.}
\label{tab:imu_error}
\centering
\small
\setlength{\tabcolsep}{10pt}
\begin{tabular}{lll}
\hline
Error component & Parameter(s) & Modeled effect \\
\hline
Sampling configuration
& $\Delta t_{\mathrm{imu}}, f_{\mathrm{imu}}$
& Integration interval and measurement rate \\

Constant bias
& $e_b,d_b$
& Constant gyroscope and accelerometer offsets \\

Random walk
& $w_{eb},w_{db}$
& Uncorrelated stochastic error accumulation \\

Correlated bias
& $R_{0G},R_{0A},\tau_G,\tau_A$
& First-order Gauss--Markov bias process \\

Scale-factor error
& $\delta K_{G,ii},\delta K_{A,ii}$
& Axis-dependent multiplicative error \\

Axis misalignment
& $\delta K_{G,ij},\delta K_{A,ij}$
& Off-diagonal cross-axis coupling \\
\hline
\end{tabular}
\end{table*}

A compact representation connects the error parameters in Table~\ref{tab:imu_error} to the sampled inertial increments:
\begin{equation}
\begin{aligned}
\Delta\tilde{\boldsymbol\theta}_k &= (I+\delta K_G)\Delta\boldsymbol\theta_k
  + [\boldsymbol e_b+\boldsymbol b_{G,k}(R_{0G},\tau_G)]\Delta t_{\mathrm{imu}}
  + \boldsymbol\epsilon_{G,k}(w_{eb}),\\
\Delta\tilde{\boldsymbol v}_k &= (I+\delta K_A)\Delta\boldsymbol v_k
  + [\boldsymbol d_b+\boldsymbol b_{A,k}(R_{0A},\tau_A)]\Delta t_{\mathrm{imu}}
  + \boldsymbol\epsilon_{A,k}(w_{db}),\qquad
f_{\mathrm{imu}}=\Delta t_{\mathrm{imu}}^{-1}.
\end{aligned}
\label{eq:imu_error_components}
\end{equation}
Here $\Delta\boldsymbol\theta_k$ and $\Delta\boldsymbol v_k$ are the ideal angular and velocity increments; $\delta K_G$ and $\delta K_A$ collect the diagonal scale-factor and off-diagonal misalignment terms. The constant offsets are $\boldsymbol e_b$ and $\boldsymbol d_b$, while $\boldsymbol b_G$ and $\boldsymbol b_A$ denote correlated-bias processes characterized by initial magnitudes $R_{0G},R_{0A}$ and correlation times $\tau_G,\tau_A$. The stochastic increment terms $\boldsymbol\epsilon_G$ and $\boldsymbol\epsilon_A$ are governed by the random-walk parameters $w_{eb}$ and $w_{db}$, respectively. The notation summarizes the configurable components without imposing an additional distributional assumption.

\subsubsection{Stereo Imaging Module}\label{sec3_2_4}

To provide metrically consistent visual observations across different airport environments, \textbf{PASS} employs a fixed forward-looking stereo camera configuration. The two virtual cameras have parallel optical axes, an image resolution of $1920\times1080$ pixels, and an optical-center baseline of $1$~m. The reference camera is mounted $0.5$~m forward of the aircraft body origin with a downward pitch of approximately $7^{\circ}$. The camera intrinsics, stereo geometry, and body-to-camera installation are held constant across all sequences. Consequently, variations in the generated imagery arise from the prescribed airport and operating conditions rather than from changes in the sensor geometry.

Rather than directly adopting the nominal rendering parameters, we calibrated the virtual cameras using a geometric procedure analogous to that employed for physical camera systems. A virtual checkerboard containing $7\times8$ squares, each measuring $0.828\times0.828$~m, was observed from multiple relative positions and orientations. The intrinsic parameters and lens-distortion coefficients of the two cameras were then estimated with Kalibr~\cite{Furgale2013Kalibr}, using the planar-pattern calibration principle introduced by Zhang~\cite{Zhang2000Calibration}. The resulting focal lengths were approximately $960$ pixels, with principal points close to the image center, while the estimated distortion coefficients were of the order of $10^{-3}$ or smaller. The mean reprojection error was approximately $0.54$ pixels, indicating subpixel consistency between the observed checkerboard features and their projections under the calibrated camera model.

\begin{figure*}[t]
    \centering
    \begin{minipage}{0.48\linewidth}
        \centering
        \includegraphics[width=\textwidth]{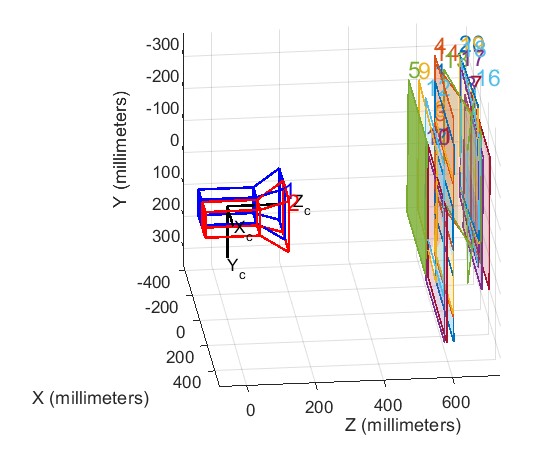}\\
        (a)
    \end{minipage}
    \hfill
    \begin{minipage}{0.48\linewidth}
        \centering
        \includegraphics[width=\textwidth]{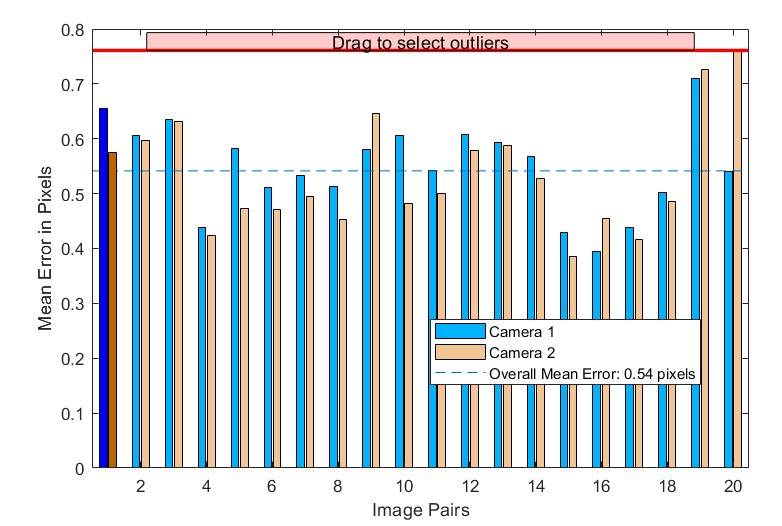}\\
        (b)
    \end{minipage}
    \caption{Virtual stereo camera calibration: (a) relative configurations of the checkerboard and camera used for multi-view calibration and (b) distribution of the resulting reprojection errors.}
    \label{fig:camera_calibration}
\end{figure*}

Figure~\ref{fig:camera_calibration} presents the calibration geometry and the distribution of reprojection residuals. The calibrated intrinsics, stereo baseline, and rigid camera-to-body transformation jointly define the metric imaging model used for stereo reconstruction and runway feature projection. Their integration into the complete coordinate transformation chain is described in Sect.~\ref{sec3_1}. Maintaining this model across all sequences separates environmental variation from calibration variation and provides a consistent visual measurement geometry for subsequent visual--inertial navigation and error analysis for downstream integrity assessment.

\begin{table}[t]
\caption{Calibrated parameters of the stereo camera system used in \textbf{PASS}. The distortion coefficients follow the order $[k_1,k_2,p_1,p_2]$.}
\label{tab:camera_calibration}
\centering
\small
\begin{tabular}{lcc}
\hline
Parameter & Left camera & Right camera \\
\hline
Image resolution (pixels)
& \multicolumn{2}{c}{$1920\times1080$} \\

$f_x$ (pixels)
& $960.6297$ & $959.6249$ \\

$f_y$ (pixels)
& $960.4265$ & $959.4180$ \\

$c_x$ (pixels)
& $962.9833$ & $961.6595$ \\

$c_y$ (pixels)
& $540.6769$ & $540.3199$ \\

$k_1$
& $1.796\times10^{-4}$ & $1.462\times10^{-3}$ \\

$k_2$
& $7.065\times10^{-4}$ & $-9.139\times10^{-4}$ \\

$p_1$
& $1.328\times10^{-4}$ & $1.233\times10^{-4}$ \\

$p_2$
& $2.798\times10^{-4}$ & $-1.289\times10^{-4}$ \\

Stereo baseline
& \multicolumn{2}{c}{$1$~m} \\

Mean reprojection error
& \multicolumn{2}{c}{$0.54$ pixels} \\
\hline
\end{tabular}
\end{table}

\subsection{Support for Navigation Integrity (CAT III)}\label{sec3_3}

Navigation integrity concerns the trust that can be placed in a navigation solution and the ability to provide a timely alert when hazardously misleading information (HMI) occurs. The development and evaluation of integrity algorithms require more than the sensor measurements used by the navigation estimator; they also require controllable error conditions, estimator-independent reference quantities, deterministic geometric priors, and sufficiently diverse samples for characterizing measurement uncertainty. A simulation system or dataset that provides these elements can support the construction, calibration, and evaluation of downstream integrity models under repeatable operating conditions. Accordingly, the integrity support capabilities of \textbf{PASS} are organized into three categories: (A) controllable scenario perturbation and fault injection, (B) independent truth and runway geometry references, and (C) condition-diverse data for statistical error modeling.
\subsubsection*{\textit{A. Controllable scenario perturbation and fault injection}}
\label{sec:controlled_scenario_perturbation}

To generate traceable error conditions, \textbf{PASS} separates the prescribed aircraft motion and reference geometry from the variables governing visual observation quality. Building on the parameterized airport and runway construction described above, illumination, weather, runway appearance, dynamic objects, and partial occlusion can be varied while the reference trajectory, nominal camera geometry, and runway coordinate frame remain unchanged. This controlled factor design allows changes in runway visibility and feature localization quality to be associated with known operating conditions rather than unintended variations in aircraft motion or reference geometry.

\begin{figure*}[t]
    \centering
    \begin{minipage}{0.8\textwidth}
        \centering
        \includegraphics[width=\linewidth]{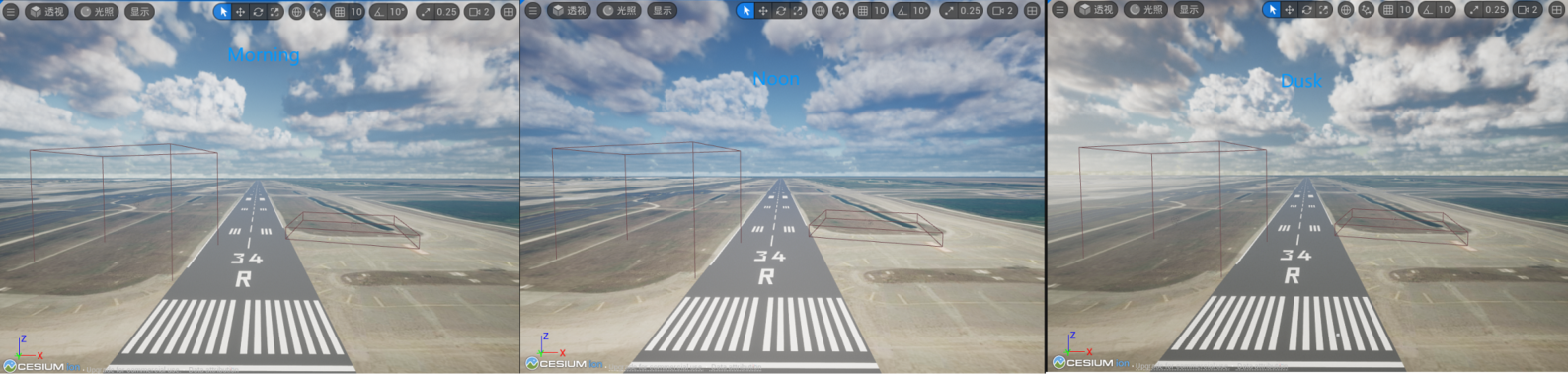}\\
        (a)
    \end{minipage}

    \begin{minipage}{0.8\textwidth}
        \centering
        \includegraphics[width=\linewidth]{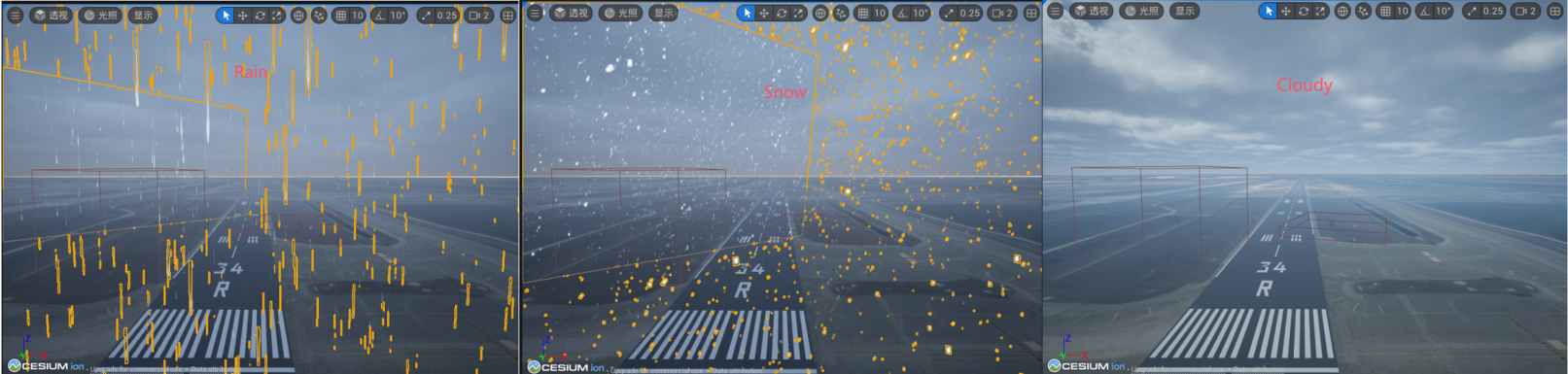}\\
        (b)
    \end{minipage}
    \caption{Controlled environmental variations represented in \textbf{PASS}: (a) illumination conditions at different times of day and (b) representative weather conditions. The prescribed aircraft trajectory and runway reference geometry remain unchanged across these variations.}
    \label{fig:environment_examples}
\end{figure*}

\begin{figure}[t]
    \centering
    \includegraphics[width=0.40\linewidth]{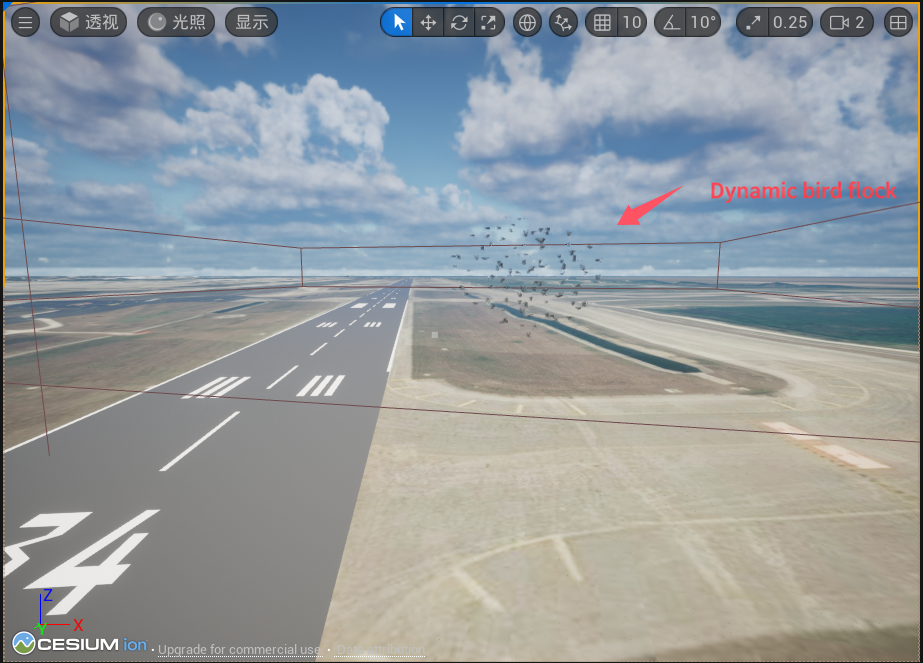}
    \caption{Dynamic-object-induced visual perturbation in the approach view. The birds introduce localized motion and partial occlusion of runway features without modifying the prescribed aircraft trajectory or reference geometry.}
    \label{fig:dynamic_objects}
\end{figure}

\textbf{PASS} further distinguishes normal scenario variation from explicit fault injection. Consistent changes to both the rendered runway and its geometric description produce additional airport or runway configurations without introducing a reference inconsistency. In contrast, perturbing the stored runway prior independently of the rendered scene produces a controlled map prior fault. Similarly, independently varying camera calibration or installation parameters can represent sensor model inconsistencies, as illustrated for visual--inertial integrity modeling in Fig.~7 of \cite{Sun2026IntegrityVINS}. These mechanisms allow downstream integrity algorithms to be evaluated using visual disturbances and prior information faults whose source and configuration are explicitly defined. Figures~\ref{fig:environment_examples} and~\ref{fig:dynamic_objects} show representative environmental variations and occlusions caused by dynamic objects generated by \textbf{PASS}.

\subsubsection*{\textit{B. Independent truth and runway geometry references}}
\label{sec:truth_geometry_references}

Integrity assessment requires reference information that is independent of the navigation solution being evaluated. \textbf{PASS} provides a continuous six-degree-of-freedom aircraft trajectory expressed in the runway frame and synchronized with the generated stereo images. Together with the calibrated camera model and the common coordinate framework introduced in Sect.~\ref{sec3_1}, this trajectory determines the ground truth camera pose at each image timestamp without relying on the output of a visual front end or navigation estimator.

\begin{figure}[t]
    \centering
    \includegraphics[width=0.60\linewidth]{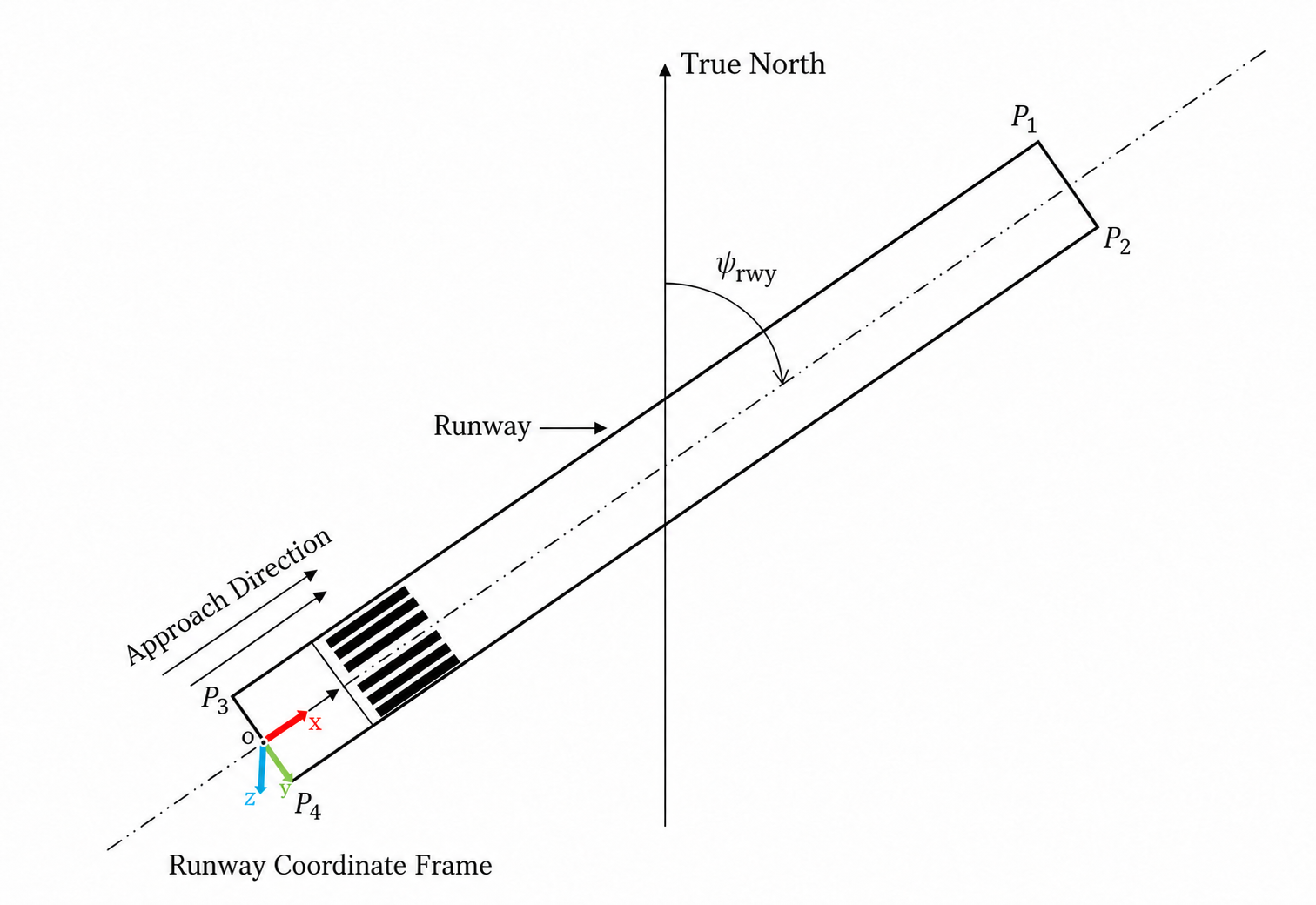}
    \caption{Runway geometry and runway-centered reference frame used by \textbf{PASS}. The runway origin, dimensions, corner coordinates, principal line features, and runway heading $\psi_{\mathrm{rwy}}$ define deterministic geometric references for runway feature projection and error evaluation. }
    \label{fig:runway_geometry}
\end{figure}

For each airport, \textbf{PASS} additionally defines the runway frame origin and axes, runway length and width, true bearing, four runway corners' coordinate, and six three-dimensional endpoints' coordinate describing the left side line, right side line, and threshold line. As illustrated in Fig.~\ref{fig:runway_geometry}, these quantities provide a deterministic geometric description of the runway rather than detector dependent image annotations.

The runway references can be transformed into the camera frames and projected onto the images using the ground truth pose and calibrated camera model. Comparing these truth-based projections with detected runway corners or lines yields a directly computable visual projection error. For line features, the error may be expressed using signed point-to-line distances between truth projected endpoints and detected image lines. Repeating this calculation throughout an approach produces a time series of truth-referenced visual error samples. \textbf{PASS} supplies the required truth and geometric references, while the feature detector, residual definition, error-overbounding method, and integrity monitoring architecture remain choices of the downstream algorithm~\cite{Sun2026IntegrityVINS}.

\subsubsection*{\textit{C. Physics-diverse data for statistical error modeling}}
\label{sec:statistical_error_modeling}

\begin{table*}[t]
\caption{Airport categories included in  \textbf{PASS} dataset.}
\label{tab:airport_geometry}
\centering
\small
\begin{tabular}{lllrrr}
\hline
Category & ICAO & Runway & Length (m) & Width (m) \\
\hline
Urban Airport & KLAX & L6--R24 & 3381.7560 & 60.9600  \\
Urban Airport & ZSPD & R34--L16 & 3799.9416 & 60.0456 \\
Desert Airport & OMAA & R31--L13 & 4099.8600 & 60.0456  \\
Mountainous Airport & LOWI & 26--08 & 2000.0976 & 45.1104  \\
Coastal Airport & LFMN & R22--L40 & 2627.9856 & 45.1104 \\
Coastal Airport & LGMK & 34--16 & 1901.9520 & 29.8704  \\
\hline
\end{tabular}
\end{table*}

Statistical error modeling requires repeated measurements collected under sufficiently diverse and identifiable operating conditions. \textbf{PASS} generates continuous approach sequences across multiple airports and runway configurations, with controlled variations in weather, runway visibility, dynamic object occlusion, and approach geometry.It is shown in Table~\ref{tab:airport_geometry} Each sequence is associated with its prescribed scenario configuration, enabling the resulting visual error samples to be grouped by airport, environmental condition, distance to the threshold, projected runway scale, or occlusion state.

Combining these condition descriptors with the truth-referenced errors supports the estimation of stage- and condition-dependent visual error distributions. The resulting samples can be used to estimate dynamic measurement covariances, construct conservative error envelopes, and examine whether a statistical error model remains applicable across different airports and operating conditions. The multi-airport configuration further helps distinguish recurring error behavior from airport specific effects and thereby supports analysis of the operating domain within which an integrity model remains trustworthy.

Taken together, the three categories establish a continuous information chain from controlled disturbance generation, through independent error evaluation, to condition-dependent statistical characterization. \textbf{PASS} thereby provides supporting information for the development and evaluation of downstream integrity algorithms without prescribing a particular navigation estimator, integrity monitoring architecture, or protection level formulation. These simulation based data do not, by themselves, establish certified CAT III integrity performance or replace validation using representative real-flight observations.

\subsection{Sim-to-Real Scene Trustworthiness inspired by Human Cognition}
\label{sec3_4}

Our objective is to assess whether simulated approach images are consistent with real images in their global organization by scene category. The practical availability of real approach imagery constrains this assessment. Access to controlled real-flight approach recordings, particularly those accompanied by camera and aircraft state metadata, was limited by confidentiality constraints. We therefore used the available real-image subset of LARD together with frames extracted from publicly available YouTube approach videos. For many of these images, the camera geometry, aircraft state, and precise approach conditions are unknown. Consequently, a real image and a simulated image assigned to the same broad airport category cannot generally be paired at the same airport and approach state. Moreover, within-category variation in airport layout, surrounding terrain, imaging conditions, and viewpoint can dominate a direct comparison of sparsely sampled high-dimensional feature distributions. These considerations do not make high-dimensional two-sample methods invalid, but they limit the interpretation of a single distributional discrepancy as evidence of scene-level correspondence in this dataset.

\begin{figure*}[t]
    \centering
    \begin{minipage}[t]{0.235\textwidth}
        \vspace{0pt}
        \centering
        \parbox[t][0.115\textheight][t]{\linewidth}{\centering\includegraphics[width=\linewidth,height=0.115\textheight,keepaspectratio]{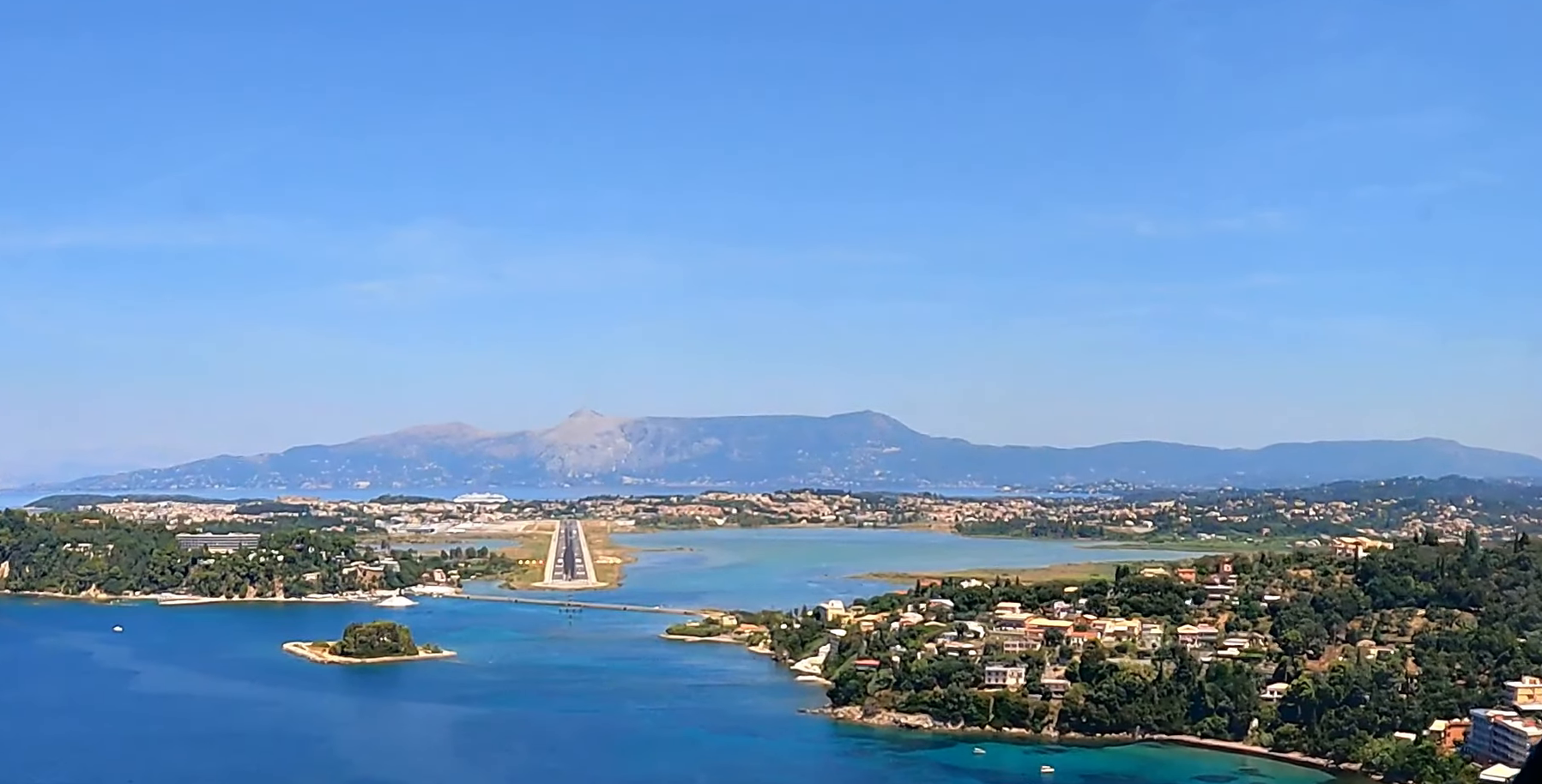}}\\[-1mm]
        {\footnotesize (a) Real coastal}
    \end{minipage}
    \hfill
    \begin{minipage}[t]{0.235\textwidth}
        \vspace{0pt}
        \centering
        \parbox[t][0.115\textheight][t]{\linewidth}{\centering\includegraphics[width=\linewidth,height=0.115\textheight,keepaspectratio]{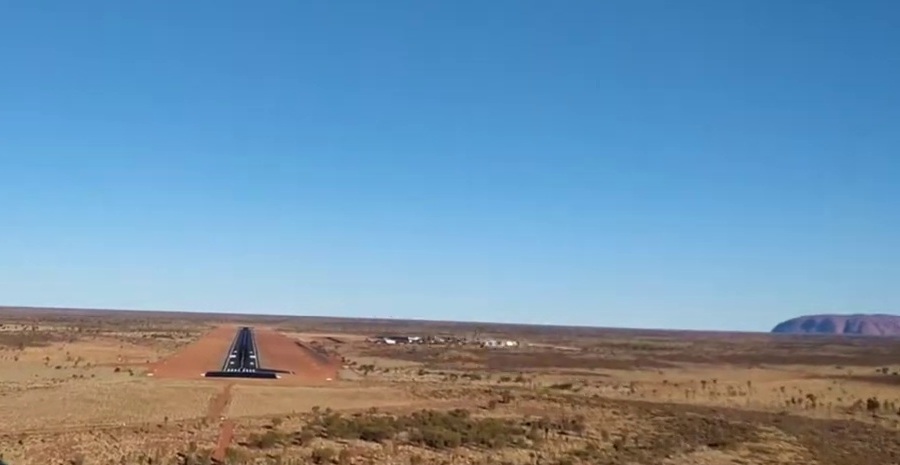}}\\[-1mm]
        {\footnotesize (b) Real desert}
    \end{minipage}
    \hfill
    \begin{minipage}[t]{0.235\textwidth}
        \vspace{0pt}
        \centering
        \parbox[t][0.115\textheight][t]{\linewidth}{\centering\includegraphics[trim=40bp 0bp 270bp 0bp,clip,width=\linewidth,height=0.115\textheight,keepaspectratio]{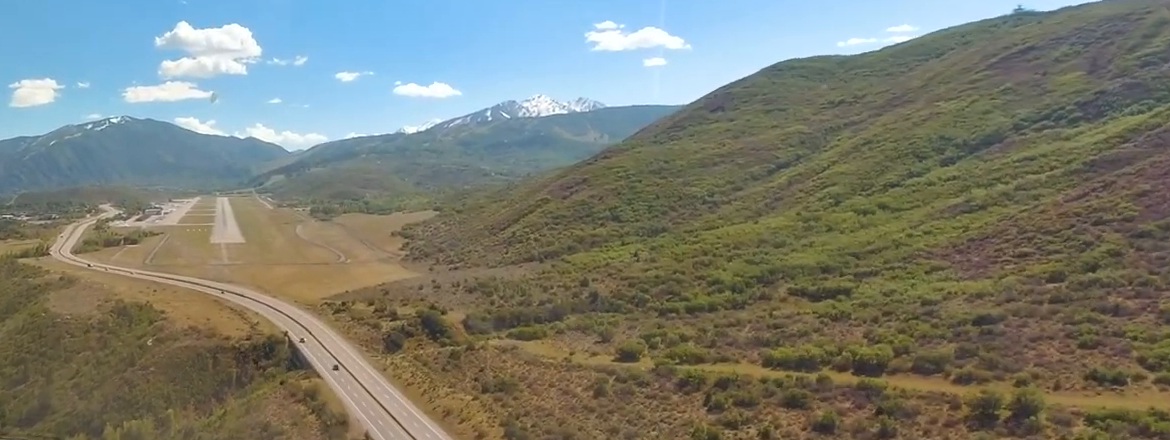}}\\[-1mm]
        {\footnotesize (c) Real mountain}
    \end{minipage}
    \hfill
    \begin{minipage}[t]{0.235\textwidth}
        \vspace{0pt}
        \centering
        \parbox[t][0.115\textheight][t]{\linewidth}{\centering\includegraphics[trim=0bp 100bp 0bp 50bp,clip,width=\linewidth,height=0.115\textheight,keepaspectratio]{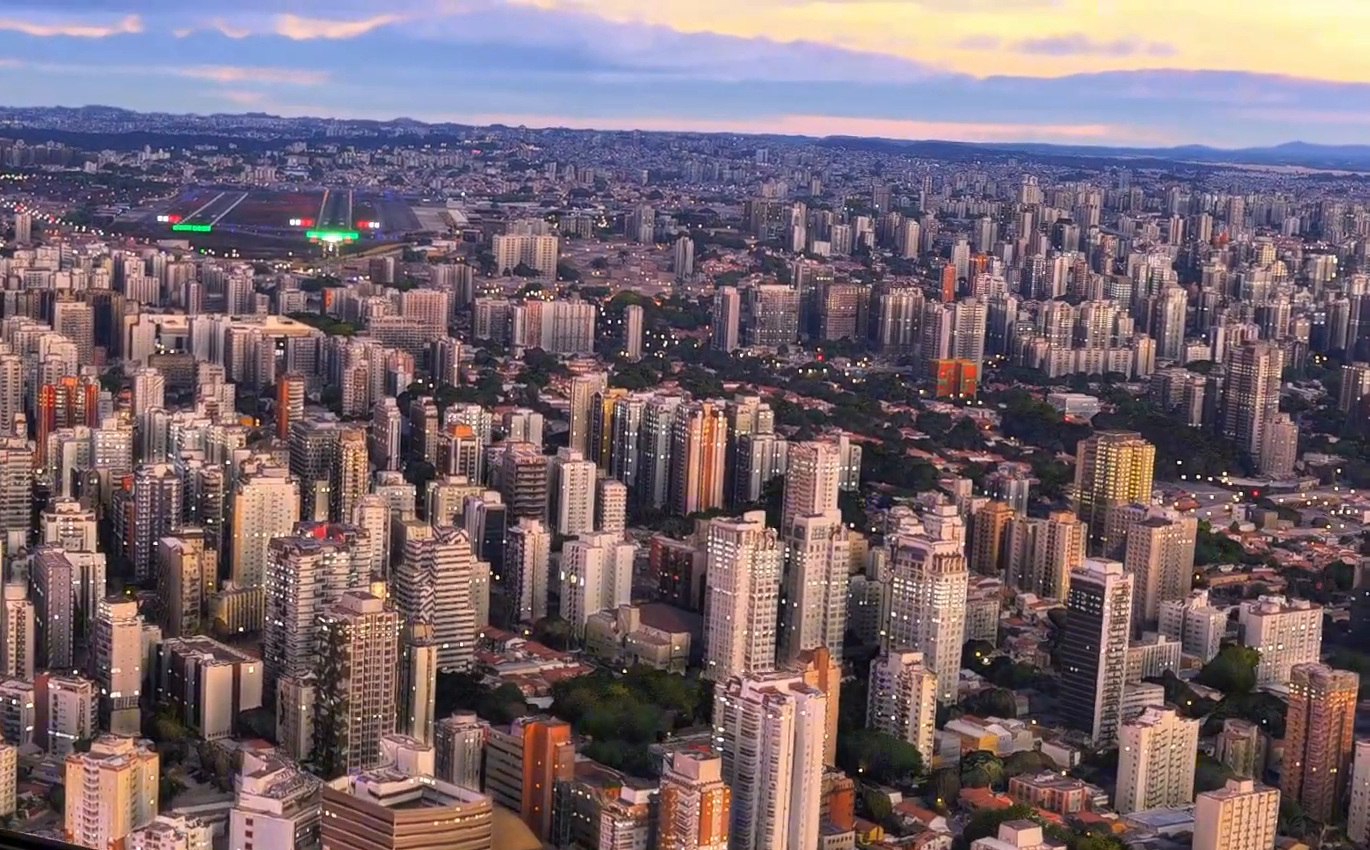}}\\[-1mm]
        {\footnotesize (d) Real urban}
    \end{minipage}

    \vspace{1.5mm}

    \begin{minipage}[t]{0.235\textwidth}
        \vspace{0pt}
        \centering
        \parbox[t][0.115\textheight][t]{\linewidth}{\centering\includegraphics[width=\linewidth,height=0.115\textheight,keepaspectratio]{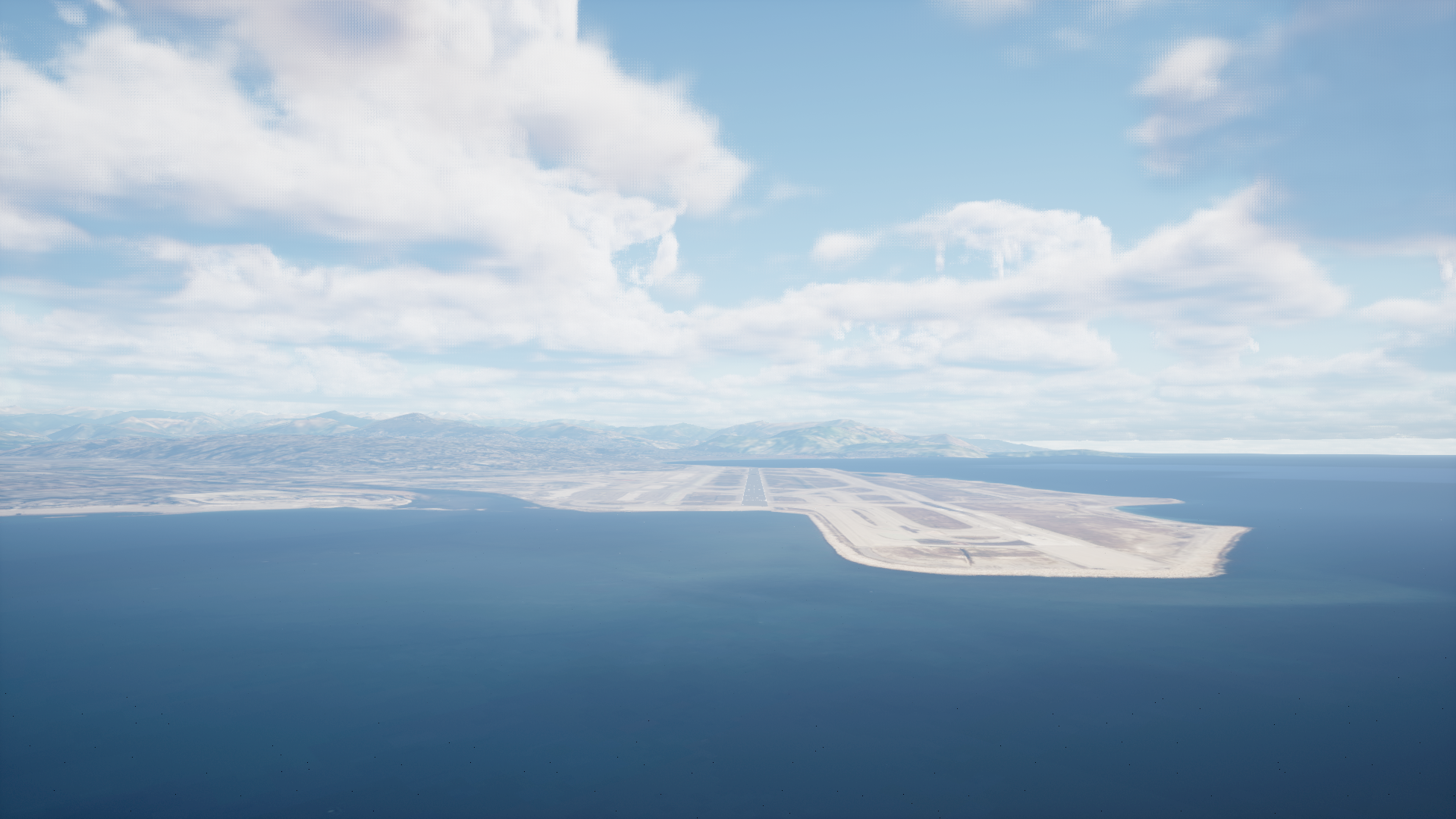}}\\[-1mm]
        {\footnotesize (e) Simulated coastal}
    \end{minipage}
    \hfill
    \begin{minipage}[t]{0.235\textwidth}
        \vspace{0pt}
        \centering
        \parbox[t][0.115\textheight][t]{\linewidth}{\centering\includegraphics[width=\linewidth,height=0.115\textheight,keepaspectratio]{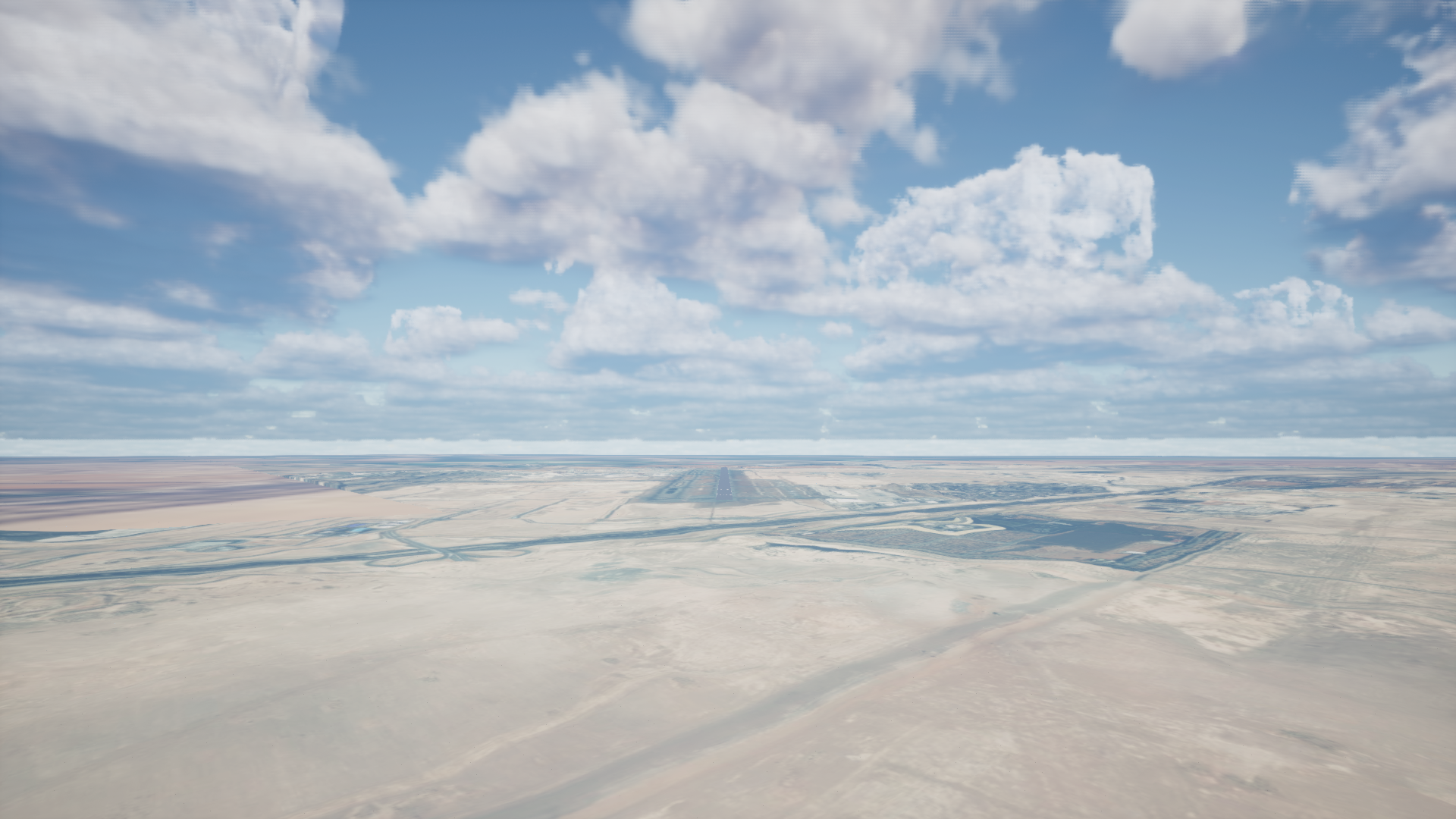}}\\[-1mm]
        {\footnotesize (f) Simulated desert}
    \end{minipage}
    \hfill
    \begin{minipage}[t]{0.235\textwidth}
        \vspace{0pt}
        \centering
        \parbox[t][0.115\textheight][t]{\linewidth}{\centering\includegraphics[width=\linewidth,height=0.115\textheight,keepaspectratio]{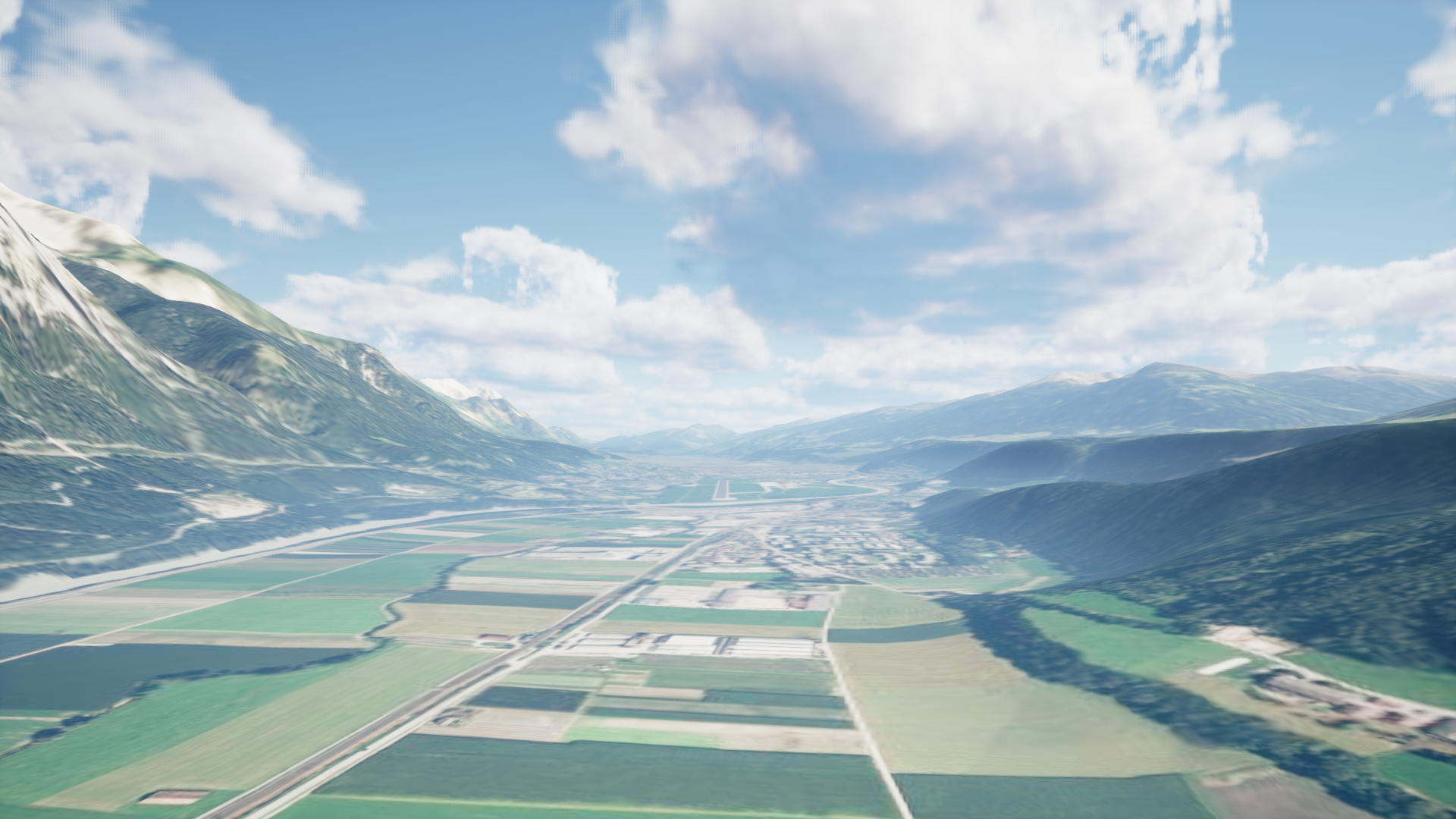}}\\[-1mm]
        {\footnotesize (g) Simulated mountain}
    \end{minipage}
    \hfill
    \begin{minipage}[t]{0.235\textwidth}
        \vspace{0pt}
        \centering
        \parbox[t][0.115\textheight][t]{\linewidth}{\centering\includegraphics[width=\linewidth,height=0.115\textheight,keepaspectratio]{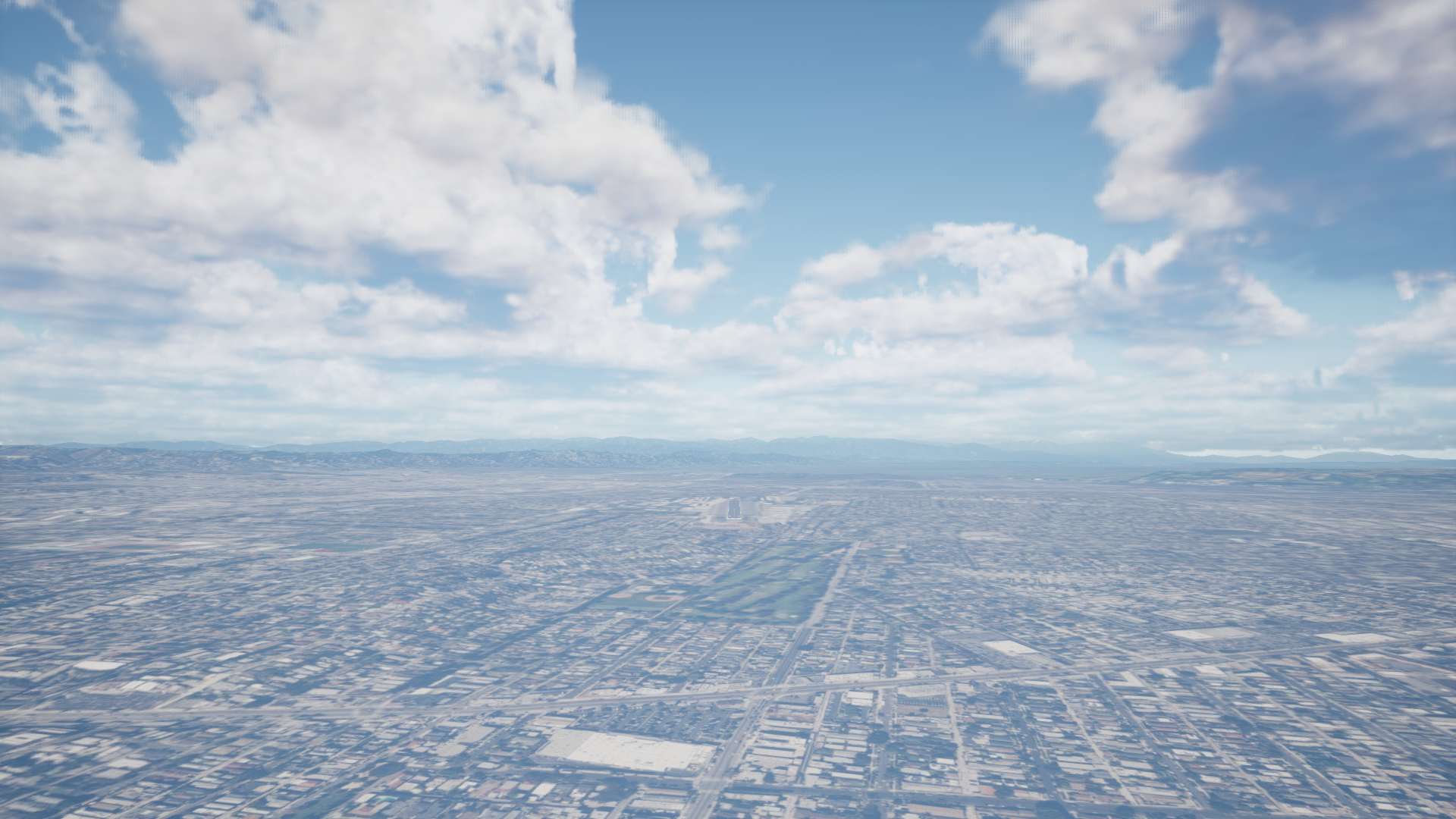}}\\[-1mm]
        {\footnotesize (h) Simulated urban}
    \end{minipage}

    \caption{Representative real and simulated approach images from coastal, desert, mountain, and urban airport environments.}
    \label{fig:sec3_4_inputs}
\end{figure*}

We define Sim-to-Real Scene Trustworthiness (SRST) as \emph{category-relative correspondence} within each approach stage. The assessment considers four representative airport environments---coastal, desert, mountainous, and urban---defined by the dominant visual context surrounding the runway. Runway scale, visible ground coverage, and scene detail evolve during approach; therefore, middle approach images, in which the runway is relatively small and more of the surrounding environment remains visible, and low approach images, in which the runway occupies more of the view, are analyzed separately to limit confounding by viewpoint evolution when comparing environmental structure across domains. Rather than requiring an exact real image counterpart, SRST examines whether a simulated image is more closely associated with its intended category in the real-image set than with competing categories. We consider this question at two category resolutions: the original four-category formulation (4c) and a three-category formulation (3c) in which coastal and urban scenes are combined. Representative real and simulated images of the four environments are shown in Fig.~\ref{fig:sec3_4_inputs}.

We use the pretrained DINOv2-small encoder~\cite{Oquab2024DINOv2} as a fixed feature extractor, without task-specific training or fine-tuning on the airport images. DINOv2 provides self-supervised visual representations that have been shown to transfer to image level tasks without downstream fine-tuning, making it a suitable task agnostic starting point for this comparison. Related work such as AnyLoc~\cite{Keetha2024AnyLoc} has used frozen self-supervised features across diverse environments without place recognition specific training. AnyLoc aggregates dense local features for place retrieval, whereas our analysis uses a single global image descriptor for category-relative scene assessment; it does not reproduce the AnyLoc aggregation or retrieval pipeline.

For each image, the class token (CLS token) from the final transformer layer is taken as a 384 dimensional global descriptor and normalized to unit $\ell_2$ norm. This image level readout matches our objective of comparing broad scene organization rather than local feature correspondences. Within each approach stage, the real and simulated descriptors are jointly embedded in two dimensions using unsupervised UMAP~\cite{McInnes2018UMAP} with cosine input distance, 50 neighbors, and a minimum distance of 0.3. Category labels are not used to fit the embedding, and two-dimensional Euclidean distances are not assumed to recover the complete geometry of the original feature space.
\newline

As a result, we propose two variants of SRST at different levels of category granularity, as follows.

\noindent\textit{Alternative 1. 4c: four-category comparison.}
Retaining coastal, desert, mountainous, and urban scenes as separate categories, we define the overall SRST as
\begin{equation}
Q^{\mathrm{SRST}}
=
\frac{1}{|\mathcal{S}_{4}|}
\sum_{s\in\mathcal{S}_{4}}
\mathbb{I}\!\left[
\sum_{j\in\mathcal{C}_{4}\setminus\{c_s\}}
\mathbb{I}\!\left(p^{(4)}_{s,j}>\frac{1}{2}\right)
\geq 2
\right],
\label{eq:srst-four-class}
\end{equation}
The resulting $Q^{\mathrm{SRST}}$ measures trustworthiness under the four-category (4c) taxonomy. Here, $\mathcal{C}_{4}$ is the set of four scene categories, $\mathcal{S}_{4}$ is the evaluated set of simulated images, and $c_s$ and $t_s$ denote the assigned category and approach stage of image $s$. Let $R^{(4)}_{t,c}$ denote the real reference images in stage $t$ and category $c$, and let $z_i\in\mathbb{R}^{2}$ be the UMAP coordinate of image $i$. For each competing category $j$, $p^{(4)}_{s,j}$ averages the comparisons between every real image $r\in R^{(4)}_{t_s,c_s}$ and every real image $r'\in R^{(4)}_{t_s,j}$. A comparison contributes one when $d(z_s,z_r)<d(z_s,z_{r'})$, one-half when the distances are equal, and zero otherwise, with $d$ denoting Euclidean distance. The indicator function $\mathbb{I}[\cdot]$ equals one when its condition is satisfied and zero otherwise. Thus, an image passes the 4c criterion when the corresponding category in the real-image set is favored over at least two of the three alternatives. A pairwise score equal to one-half does not constitute a category win.
\newline

\noindent\textit{Alternative 2. 3c: merged coastal--urban comparison.}
Coastal and urban approaches need not form mutually exclusive visual
environments: buildings and roads may appear beside a coastal runway,
while water or shoreline may be visible in an urban approach. To assess
scene correspondence at a coarser category resolution, we additionally
combine these environments into one urban/coastal category and retain
desert and mountainous scenes separately. The corresponding overall
SRST is
\begin{equation}
Q^{\mathrm{SRST}}_{\mathrm{3c}}
=
\frac{1}{|\mathcal{S}_{3}|}
\sum_{s\in\mathcal{S}_{3}}
\mathbb{I}\!\left[
\min_{j\in\mathcal{C}_{3}\setminus\{c_s\}}
p^{(3)}_{s,j}>\frac{1}{2}
\right],
\label{eq:srst-three-class}
\end{equation}
The resulting $Q^{\mathrm{SRST}}$ measures trustworthiness under the three-category (3c) taxonomy. Here, $\mathcal{C}_{3}$ comprises urban/coastal, desert, and mountainous scenes, and $\mathcal{S}_{3}$ contains the simulated images assigned to these categories for this assessment. The score $p^{(3)}_{s,j}$ follows the same all pairs comparison as $p^{(4)}_{s,j}$, but is recalculated after pooling the real coastal and urban reference images. Since two competing categories remain, Eq.~\eqref{eq:srst-three-class} requires the assigned category to be favored over both of them. The 4c and 3c evaluations thus assess scene correspondence at different category resolutions and should not be interpreted as interchangeable estimates of the same classification task.

\begin{figure}[t]
    \centering
    \includegraphics[width=1.0\linewidth]{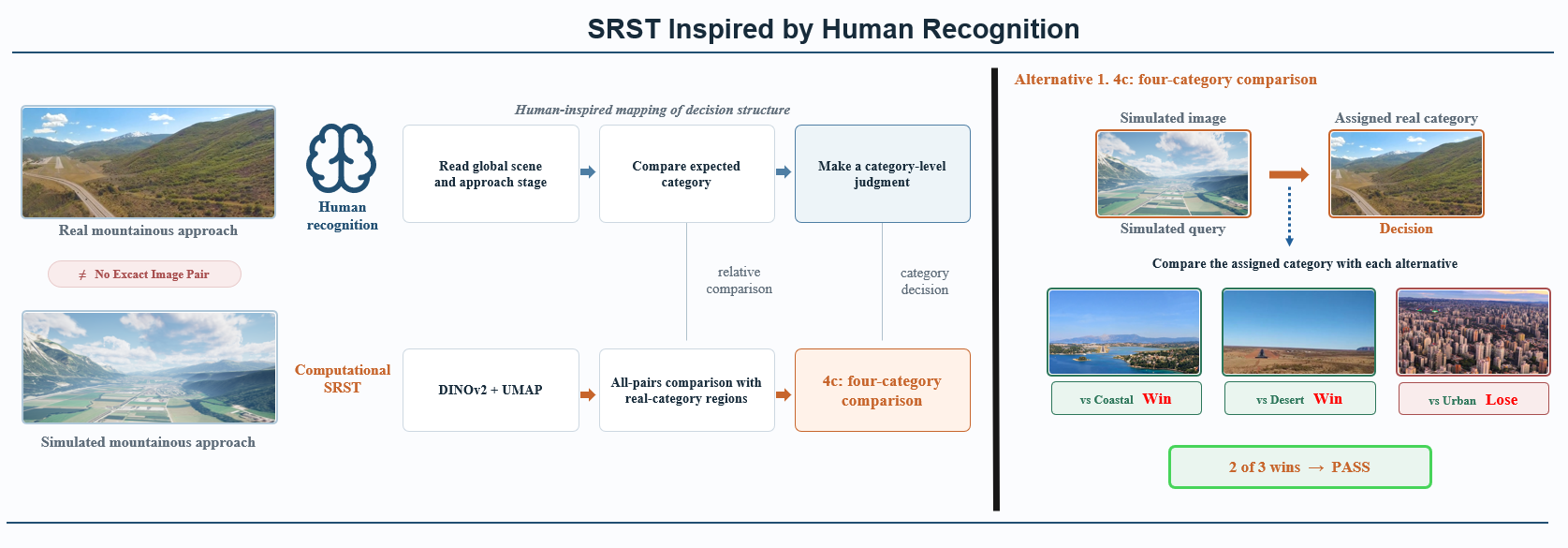}
    \caption{Human-inspired decision process of SRST for unpaired real and simulated approach images. The 4c formulation requires the assigned category to prevail over at least two of three alternatives, whereas the 3c formulation merges coastal and urban scenes and requires the assigned category to prevail over both remaining alternatives.}
    \label{fig:Human-recognition}
\end{figure}

The rationale behind the two SRST formulations is that, in the absence of an exactly paired real image, the trustworthiness of a simulated view can be judged from whether it conveys a coherent scene identity relative to reasonable alternatives. Human observers naturally prioritize information that is important for understanding the overall scene, rather than relying only on locally salient content \cite{Murlidaran2026SceneUnderstanding}. Such scene-level interpretation is consistent with cognitive models in which natural images are encoded into high dimensional visual representations and categorized according to their similarity to candidate prototypes or exemplars \cite{Battleday2020HumanCategorization}. When several interpretations are plausible, the competing categories also provide an informative reference: human confidence in multi-alternative decisions has been shown to depend strongly on the evidence difference between the leading and next best alternatives \cite{Li2020MultiAlternativeConfidence}. SRST translates these insights into a category relative assessment of simulated airport approach imagery. It reads the global scene in the context of the approach stage, evaluates whether the expected category is favored over plausible alternatives, and produces a category-level judgment of perceptual fidelity, as illustrated in Fig.~\ref{fig:Human-recognition}.

The comparative structure of SRST is intended to approximate one aspect of human scene judgment. When an exact reference view is unavailable, an observer may identify a scene by considering whether its overall visual context is more consistent with the expected category than with plausible alternatives. SRST expresses this reasoning through comparisons with real-category regions rather than through a single nearest image: 4c requires the assigned category to prevail over most alternatives, whereas 3c requires it to prevail over both alternatives under a coarser taxonomy. This is an interpretable analogy to a category-level decision, not an empirical demonstration that the metric reproduces human judgments. In the present study, a score exceeding 0.90 serves as an operational criterion for favorable category relative correspondence in the evaluated images, rather than an aviation safety threshold.

\section{Experiments}\label{sec4}

The experiments examine two complementary properties of \textbf{PASS}. The integrity support information provided by \textbf{PASS} warrants a dedicated evaluation. Sun et al.~\cite{Sun2026IntegrityVINS} have used \textbf{PASS} to evaluate a visual--inertial navigation integrity method; that analysis is therefore not repeated here. Their method employs absolute localization, whereas the present study uses open-source stereo--inertial estimators to assess the functional usability of the dataset. Because the algorithms and evaluation objectives differ, their reported positioning errors should not be treated as directly comparable measures of localization accuracy. Section~\ref{sec4_1} evaluates the functional usability of the generated stereo--inertial sequences by determining whether representative state estimators can consume the released sensor streams and produce quantitatively evaluable trajectories during terminal approach. Section~\ref{sec4_2} then assesses the sim-to-real trustworthiness of the visual observations using the feature-space methodology introduced in Sect.~\ref{sec3_4}. Together, these experiments address both the internal consistency of the generated data and the correspondence between simulated and real approach imagery.

\subsection{Functional Usability of \textbf{PASS} Sequences}
\label{sec4_1}

Airborne visual odometry becomes increasingly difficult when the observed scene is distant relative to the stereo baseline. Under this geometry, small image disparities weaken depth estimation and make motion recovery sensitive to feature localization and calibration errors~\cite{Warren2013HighAltitude}. During high approach, when the aircraft is farther from the runway and distant scene structure dominates the view, visual--inertial estimation is further affected by limited parallax and by the numerical sensitivity of estimator initialization~\cite{Liu2017HighAltitudeVINS}. These difficulties are particularly relevant to fixed-wing approach, where the runway and its markings initially occupy only a small portion of the image and much of the visible ground structure is approximately planar. The objective of this experiment is therefore not to establish a performance ranking between estimators or to claim compliance with an aircraft navigation accuracy requirement. Instead, we examine whether the images, inertial measurements, calibration parameters, timestamps, and runway-relative reference states generated by \textbf{PASS} constitute a coherent interface for representative visual--inertial estimation, and whether the resulting trajectories expose measurable terminal error and drift.

To separate this usability question from the distinct problem of high-altitude estimator initialization, each evaluation window was supplied with a one-time initial navigation state. This setting is consistent with a staged precision approach architecture in which the navigation solution available before the terminal segment provides an initial position, attitude, and velocity for subsequent visual--inertial estimation. In the experiment, position and attitude were assigned at the first image, while the initial velocity was fitted from the preceding $0.3$~s of reference motion. No further ground-truth state, reset, or prior update was introduced after initialization. The supplied state was ideal and is used here only to control the initial reference; robustness to uncertainty in the initial navigation solution remains outside the scope of this experiment.

We evaluated the Shanghai Pudong International Airport (ZSPD), Santa Monica Municipal Airport (KSMO), and Mykonos Airport (LGMK) sequences using VINS-Fusion~\cite{Qin2019VINSFusion} and ORB-SLAM3~\cite{Campos2021ORBSLAM3} in stereo--inertial mode. Four terminal windows of $15$, $20$, $25$, and $30$~s were extracted from each sequence, all ending at the same terminal frame. The longer windows consequently begin at an earlier approach stage and also increase the duration over which estimation errors can accumulate. Differences among window lengths should therefore be interpreted as the combined effect of approach stage and estimation interval, rather than as an isolated effect of altitude. Each airport--estimator--window combination was executed three times in an independent process, yielding $72$ stereo--inertial runs in total. These repetitions characterize estimator repeatability under identical inputs and configurations; they do not represent independent flight realizations.

The evaluation focused on whether each estimator produced a continuous trajectory over the prescribed terminal window and on the errors observed relative to the runway frame reference. Terminal position error was defined as the Euclidean distance between the estimated and reference positions at the final planned frame, while the temporal evolution of the position error was used to characterize drift within each window. A terminal error was reported only when the estimator reached the planned final frame; interrupted trajectories were retained for failure analysis but were excluded from the endpoint statistics.

The prescribed initial state was successfully applied in all runs. VINS-Fusion reached the planned terminal frame for every airport and window, whereas ORB-SLAM3 produced complete trajectories in all but one repetition of the $30$~s KSMO window. The latter terminated during backend bundle adjustment after producing a partial trajectory. This isolated interruption was retained as an observed estimator--sequence interaction rather than removed from the analysis. Overall, both estimators generated continuous and quantitatively evaluable trajectories from the stereo--inertial measurements provided by \textbf{PASS} across the three airport environments.

Table~\ref{tab:sec4_endpoint_error} reports the mean terminal position error over completed repetitions. For both estimators and at every airport, the error increased monotonically as the evaluation window was extended. At $30$~s, the mean terminal errors of VINS-Fusion were $19.95$, $40.51$, and $23.74$~m for ZSPD, KSMO, and LGMK, respectively; the corresponding ORB-SLAM3 errors were $3.02$, $18.51$, and $5.48$~m. KSMO produced the largest terminal error for both estimators throughout the four windows. Because the algorithms differ in their feature processing, initialization, state representation, and optimization strategy, these values are not used to identify a preferred estimator. Rather, their dependence on both airport and window length shows that \textbf{PASS} provides non-trivial sequences in which estimator behavior can be examined under a common sensor and ground-truth interface.

\begin{table*}[t]
\caption{Mean terminal position error [m] for the stereo--inertial terminal approach experiments. Values are in metres and are averaged over the repetitions that reached the planned final frame. Each condition contains three runs unless marked otherwise.}
\label{tab:sec4_endpoint_error}
\centering
\small
\begin{tabular}{llcccc}
\toprule
Airport & Estimator & $15$~s & $20$~s & $25$~s & $30$~s \\
\midrule
ZSPD & VINS-Fusion & 3.66  & 7.48  & 12.65 & 19.95 \\
     & ORB-SLAM3   & 0.78  & 1.32  & 2.02  & 3.02  \\
KSMO & VINS-Fusion & 12.45 & 20.18 & 30.58 & 40.51 \\
     & ORB-SLAM3   & 7.50  & 10.66 & 14.15 & 18.51$^{\dagger}$ \\
LGMK & VINS-Fusion & 9.64  & 13.87 & 18.40 & 23.74 \\
     & ORB-SLAM3   & 1.40  & 2.41  & 3.81  & 5.48  \\
\bottomrule
\end{tabular}

\vspace{2pt}
\parbox{0.96\textwidth}{
\footnotesize
$^{\dagger}$Mean over the two repetitions that reached the planned endpoint; the third repetition produced a partial trajectory and was excluded from the endpoint mean.
}
\end{table*}

\begin{figure*}[t]
    \centering
    \includegraphics[width=0.96\textwidth]
    {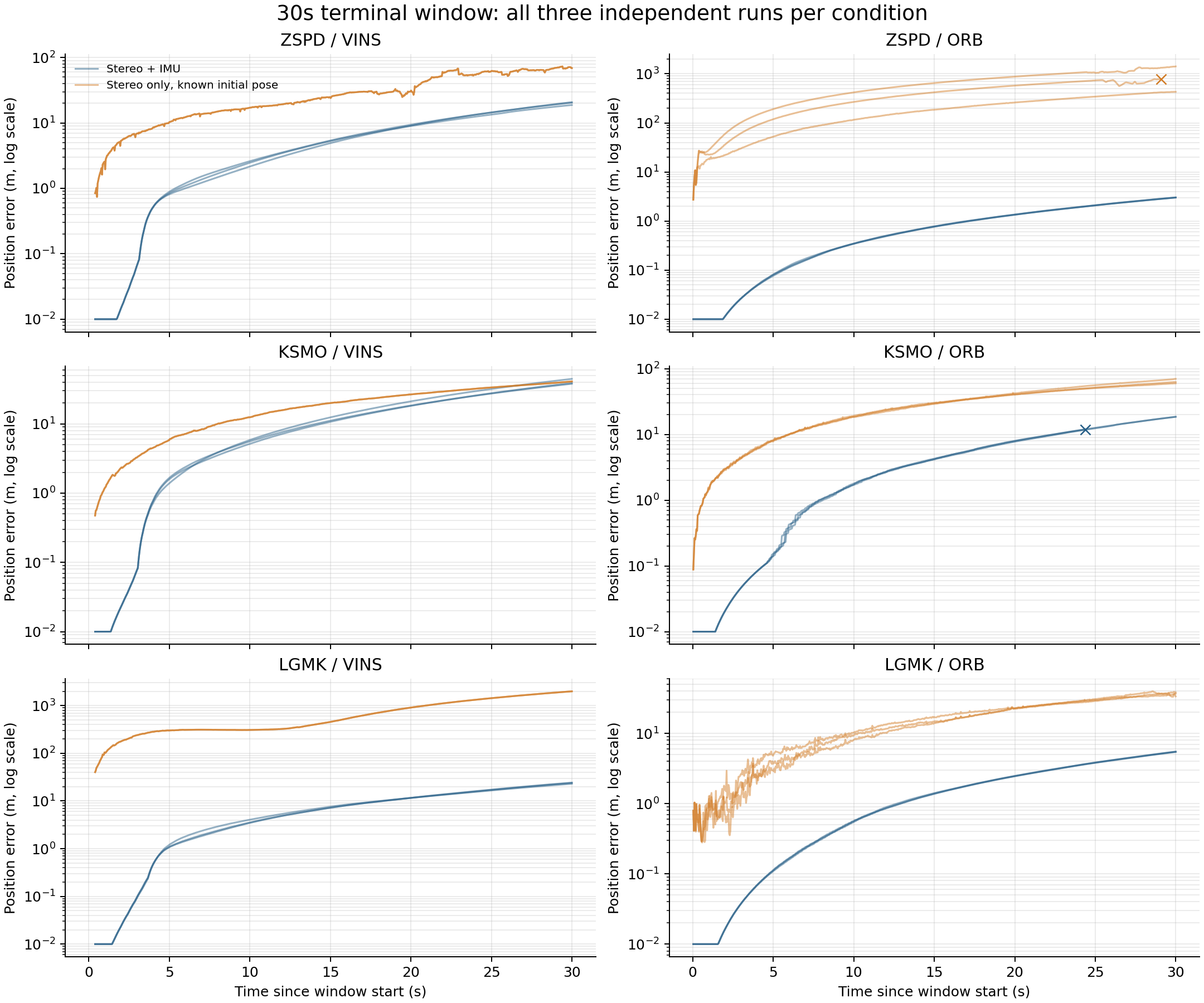}
    \caption{Evolution of the runway-relative position error over the $30$~s terminal approach windows. Each panel shows three independent executions under an identical input and configuration. The cross denotes the last valid output of the interrupted KSMO ORB-SLAM3 run and does not represent the planned terminal error. The vertical axes use a logarithmic scale.}
    \label{fig:sec4_stereo_imu_drift}
\end{figure*}

The time-resolved errors in Fig.~\ref{fig:sec4_stereo_imu_drift} provide a consistent view of drift accumulation. The three repetitions of each completed condition remain closely grouped, indicating that the principal error pattern was associated with the fixed estimator--sequence combination rather than run-to-run variation. At the same time, the growth rates differed markedly across airports and estimators. The single interrupted KSMO ORB-SLAM3 run is indicated at its final valid output and is not interpreted as a terminal measurement. These observations show that extending the estimation interval exposes progressively larger accumulated errors, while their magnitude remains conditional on both the visual environment and the estimator. Since the longer windows also start earlier in the approach, the observed trend cannot be attributed to duration or altitude alone.

The experiment therefore establishes functional usability in a deliberately bounded sense: \textbf{PASS} supplies temporally and geometrically consistent stereo--inertial observations that can drive representative estimators and support error evaluation against a runway-relative reference. It does not demonstrate that either estimator satisfies a precision approach accuracy or integrity requirement. Instead, the estimator-dependent and environment-dependent error sequences obtained here provide observable samples for subsequent statistical error characterization and downstream navigation integrity modeling.

\subsection{Sim-to-Real trustworthiness of \textbf{PASS} Sequences}
\label{sec4_2}

The real-image collection combined the real-footage subset of LARD dataset~\cite{Ducoffe2023LARD} with frames extracted from publicly available YouTube approach videos. Approximately 80 candidate videos were collected for screening. Despite the volume of available footage, many frames were unsuitable because of limited image clarity, inadequate runway visibility, or content outside the relevant approach view, including cockpit structures obscuring the scene. After screening, temporal frame selection, and cropping of extraneous regions, the two sources yielded 475 real images. Of these, 469 middle- and low-approach images were evaluated here; six touchdown images fell outside the two analyzed stages.

The simulated pool comprised 404 representative approach images from our dataset. To avoid selecting consecutive frames from a trajectory, images were sampled at approximately 4 seconds intervals and retained when they provided representative views of the four airport environments. Table~\ref{tab:cohort} reports the stage- and category-conditioned composition of both domains. Different aircraft approach phases and different categories of airports display varying levels of information in approach images. The available real footage did not consistently provide the complete airport, viewing-geometry, aircraft-state, and environmental information needed to match each real frame to a simulated image under equivalent approach conditions. The resulting images were therefore unpaired; this constraint motivates the category-relative SRST assessment rather than a direct matched-scene similarity test.

\begin{table}[htbp]
\centering
\caption{Real and simulated image counts by approach stage and airport scene category.}
\label{tab:cohort}
\begin{tabular}{lrrrr}
\toprule
& \multicolumn{2}{c}{Middle approach} & \multicolumn{2}{c}{Low approach}\\
\cmidrule(lr){2-3}\cmidrule(lr){4-5}
Category & Real & Simulated & Real & Simulated\\
\midrule
Coastal & 89 & 41 & 62 & 60\\
Desert & 27 & 61 & 34 & 40\\
Mountainous & 56 & 59 & 52 & 42\\
Urban & 104 & 79 & 45 & 22\\
\bottomrule
\end{tabular}
\end{table}

\begin{figure*}[t]
    \centering
    \includegraphics[width=\textwidth]{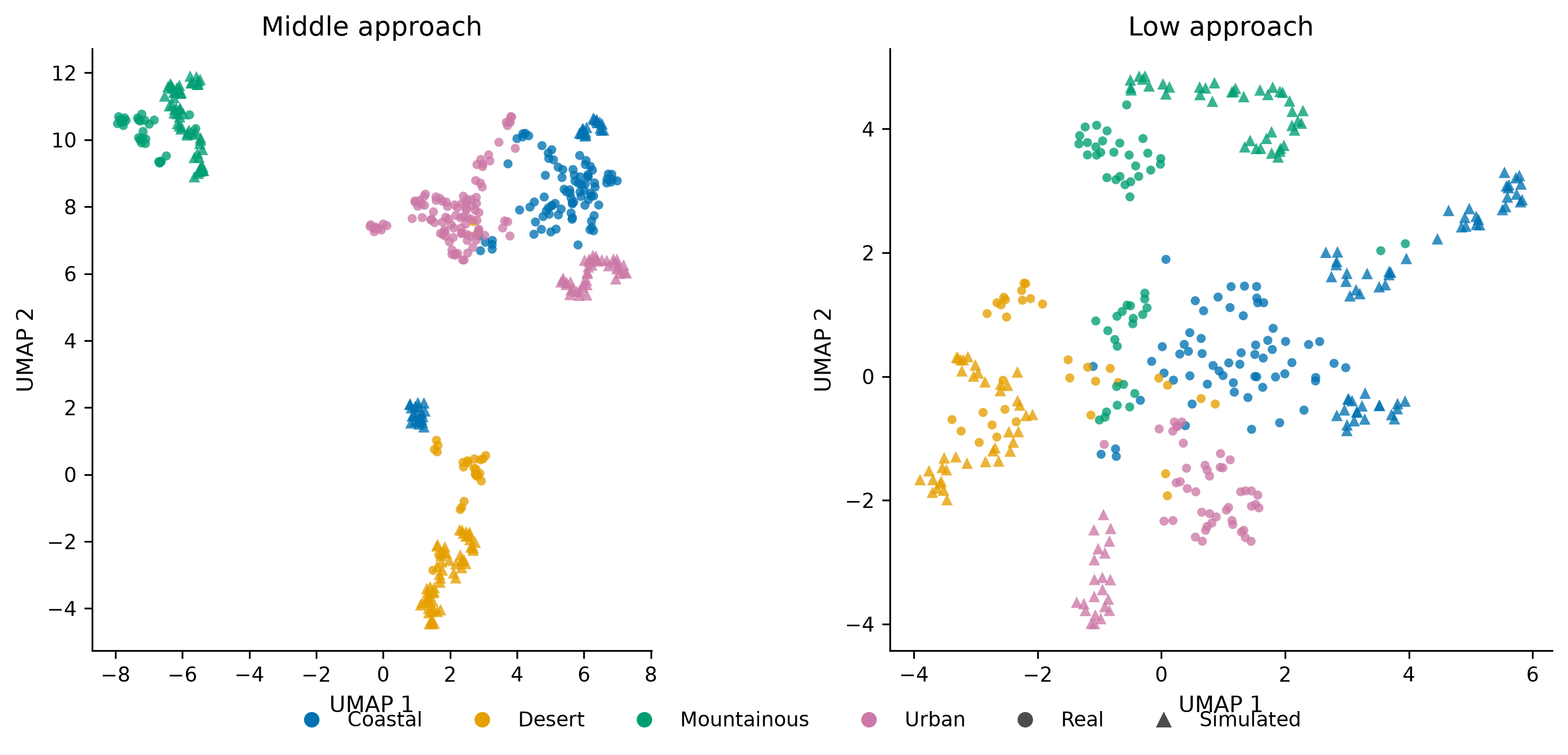}
    \caption{Joint two-dimensional UMAP embeddings of 469 real and 404 simulated approach images, fitted separately for the middle and low stages of approach. Colors indicate airport scene categories; circles and triangles denote real and simulated images, respectively. Coordinates are comparable within, but not between, the two panels. }
    \label{fig:srst_joint_umap}
\end{figure*}

\noindent\textit{Experiment 1. 4c: four-category scene correspondence.}
Figure~\ref{fig:srst_joint_umap} displays the joint UMAP embedding of the evaluated real and simulated images separately for the middle and low stages. The category colors show related scene organization across the two domains, while the marker shapes reveal that real and simulated samples can occupy adjacent but distinct subclusters. 

In the 4c assessment, a simulated image satisfies the criterion when its intended category in the real-image set is favored over at least two of the three alternatives. In the five reported realizations, 364--383 of the 404 simulated images (90.1--94.8\%) met this criterion (Table~\ref{tab:selected}). Under the operational $Q^{\mathrm{SRST}}>0.90$ criterion, most evaluated simulated images therefore exhibited correspondence with their intended real scene categories in these realizations. All images in the middle-stage desert and mountainous groups and all four low-stage groups satisfied the criterion in the five reported realizations. The remaining discrepancies were concentrated in the middle-stage coastal and urban categories.

\begin{table}[htbp]
\centering
\caption{SRST results for five selected UMAP realizations. \textbf{The six unlisted stage-by-category groups each had a 100\% image-level concordance rate in all five realizations.} }
\label{tab:selected}
\begin{tabular}{rrrr}
\toprule
UMAP Test & Overall, $n/404$ & Middle coastal, $n/41$ & Middle urban, $n/79$\\
\midrule
Test 1 & 364 (90.1\%) & 20 (48.8\%) & 60 (75.9\%)\\
Test 2 & 383 (94.8\%) & 20 (48.8\%) & 79 (100.0\%)\\
Test 3 & 365 (90.3\%) & 38 (92.7\%) & 43 (54.4\%)\\
Test 4 & 368 (91.1\%) & 20 (48.8\%) & 64 (81.0\%)\\
Test 5 & 366 (90.6\%) & 20 (48.8\%) & 62 (78.5\%)\\
\bottomrule
\end{tabular}
\end{table}

In Fig.~\ref{fig:srst_joint_umap}, the middle approach, mountainous images form a largely separated region, and desert images occupy a distinct lower region. Coastal and urban images, by contrast, lie in neighboring areas, with some real-image points interspersed near their boundary. Their proximity is consistent with the shared visual components of these environments: coastal approaches may include buildings and roads, while urban approaches may include water or shoreline. In low approach, the categories remain recognizable, but their real-image neighborhoods are less isolated and several colors meet near the central part of the embedding. Taken together, the panels suggest a qualitative shift from more clearly separated environmental contexts in middle approach to greater overlap of scene appearances in low approach. The increasing portion of the view occupied by the runway is a plausible explanation for the reduced prominence of surrounding context, but neither runway occupancy nor feature attribution was measured here. 

The principal category-level ambiguity in the 4c assessment concerns coastal and urban scenes. Their labels describe the dominant surrounding context rather than mutually exclusive visual components: coastal views may contain buildings and roads, while urban views may include water or shoreline. Such shared elements could make the two scene representations less distinct. By contrast, the arid surfaces and mountainous relief visible in the sampled desert and mountainous scenes provide more distinctive contextual cues, consistent with their results in the selected realizations. These are interpretations of scene composition; the available unpaired images cannot separate the effects of background, viewpoint, and approach conditions. The coastal--urban overlap nevertheless motivates assessing whether the same images show category-relative correspondence when this shared visual context is represented by one category.

The coastal--urban overlap observed in the four-category analysis
motivated a complementary assessment at the coarser resolution defined
by Eq.~\eqref{eq:srst-three-class}. This assessment did not alter the
image descriptors or the UMAP fitting procedure. It changed the scene
taxonomy and, for the reported exploratory cohort, the images eligible
for scoring.   61 simulated images
were assigned to a mixed arid--coastal scene and were not scored under
the three-category taxonomy. The resulting assessment used 444 real
reference images and 307 simulated images; all 469 real and 404
simulated images nevertheless remained in each stage-specific UMAP fit.
\newline

\noindent\textit{Experiment 2. 3c: merged coastal--urban scene correspondence.}
The 3c assessment combined coastal and urban scenes according to Eq.~\eqref{eq:srst-three-class}, without changing the image descriptors or the stage-specific UMAP fitting procedure. Its scoring set retained the earlier exploratory visualization mask; 61 simulated images assigned to a mixed arid--coastal scene were also left outside the three-category taxonomy. The resulting assessment used 444 real reference images and 307 simulated images, whereas all 469 real and 404 simulated images remained in each stage-specific UMAP fit. These exclusions define the scope of the reported score and were not derived from a prospectively validated image-quality rule.

Across 50 UMAP random initializations, $Q^{\mathrm{SRST}}_{\mathrm{3c}}$ averaged 98.08\%, with a between-realization standard deviation of 2.72 percentage points and a range of 90.55--100\%. All 50 realizations exceeded the descriptive 90\% criterion. Table~\ref{tab:srst-3c} presents five illustrative high-scoring realizations together with the overall mean across all 50. The selected rows illustrate the category comparisons.

\begin{table}[htbp]
\centering
\caption{Three-category SRST results. The five illustrated realizations
were selected for their high overall scores and do not estimate
performance across random initializations. The final row reports the
mean across all 50 realizations. The four unlisted stage-by-category
groups passed completely in the five illustrated realizations.}
\label{tab:srst-3c}
\begin{tabular}{lrrr}
\toprule
UMAP test & Overall, $n/307$
& Middle urban/coastal, $n/63$
& Low urban/coastal, $n/42$\\
\midrule
Test 1 & 307 (100.0\%) & 63 (100.0\%) & 42 (100.0\%)\\
Test 2 & 306 (99.7\%)  & 63 (100.0\%) & 41 (97.6\%)\\
Test 3 & 305 (99.3\%)  & 63 (100.0\%) & 40 (95.2\%)\\
Test 4 & 304 (99.0\%)  & 63 (100.0\%) & 39 (92.9\%)\\
Test 5 & 303 (98.7\%)  & 63 (100.0\%) & 38 (90.5\%)\\
\midrule
Mean of 50 tests & 98.1\% &  & \\
\bottomrule
\end{tabular}
\end{table}

\begin{figure*}[t]
    \centering
    \includegraphics[width=\textwidth]
    {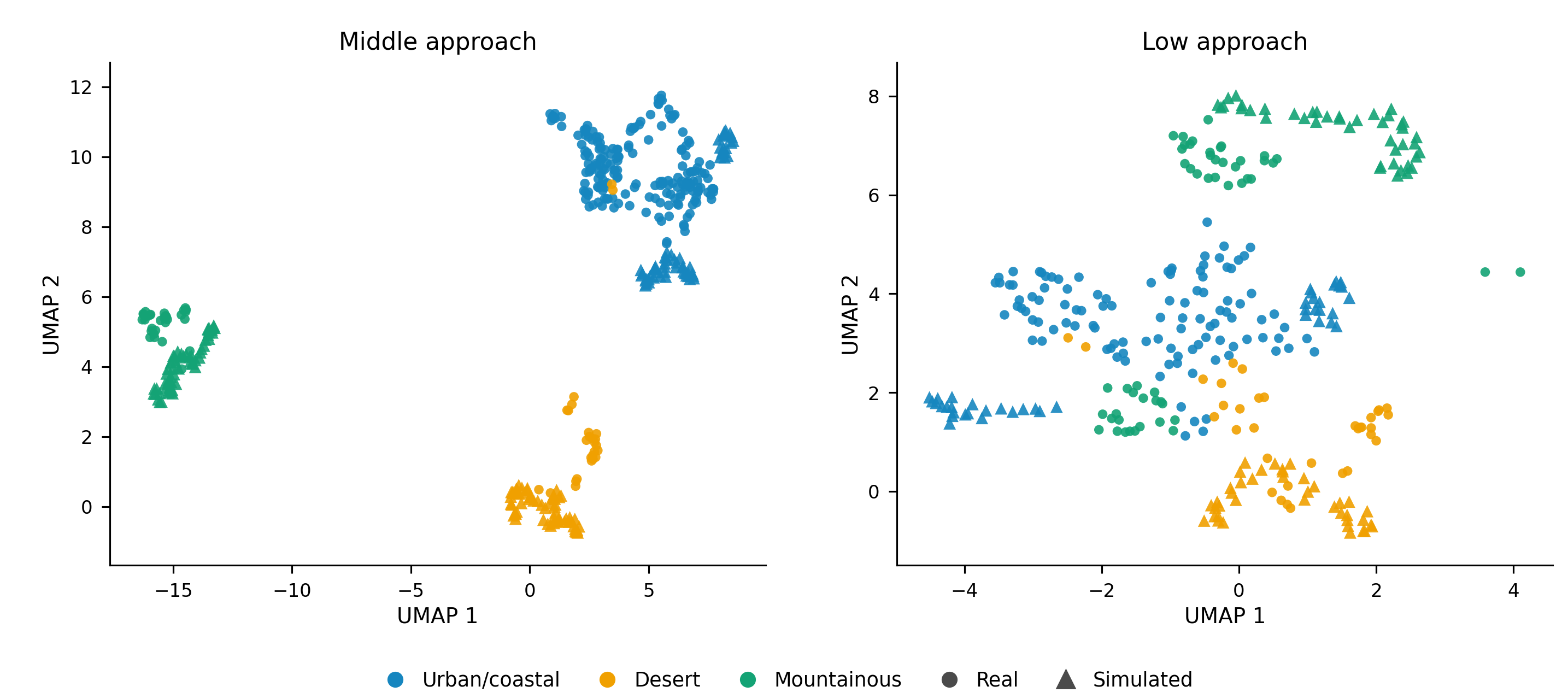}
    \caption{Joint two-dimensional UMAP embeddings for one illustrative
    three-category realization with an overall SRST score of 98.7\%.
    Middle and low approach are shown separately. Colors indicate the
    merged urban/coastal, desert, and mountainous scene categories;
    circles and triangles denote real and simulated images,
    respectively.}
    \label{fig:srst-three-class-umap}
\end{figure*}

Figure~\ref{fig:srst-three-class-umap} illustrates how the scene organization is interpreted under the broader 3c taxonomy. In middle approach, the combined urban/coastal images occupy a broad region apart from the mountainous and desert groups. This arrangement is consistent with the neighboring coastal and urban regions observed in the 4c visualization: the 3c formulation treats their shared visual context as one category rather than requiring a distinction between them. In low approach, the mountainous images remain predominantly in the upper part of the panel and desert images in the lower part, while the urban/coastal category extends across a broader central area. The boundaries between these neighborhoods are less clear than in middle approach, yet their category-level organization remains visible and most scored simulated images satisfy the 3c criterion in this realization. Thus, the change from 4c to 3c addresses an ambiguity in the \emph{category definition}, rather than demonstrating that relabeling brought real and simulated feature points closer together.  

In conclusion, the two formulations support category-relative scene correspondence for the images evaluated under their respective category definitions. They also show why the choice of scene taxonomy matters: within this collection, urban scenes are less well represented as a category visually exclusive of coastal scenes than desert or mountainous scenes. Merging coastal and urban views provides a more coherent description of their shared context.

\section{Conclusion}\label{sec5}

This paper presented \textbf{PASS}, a trustworthy synthetic stereo--inertial dataset and extensible simulation framework for fixed-wing runway approach navigation and integrity algorithm development. By coupling PSINS-based trajectory and inertial simulation with multi-airport Unreal Engine environments, \textbf{PASS} provides synchronized stereo images and IMU measurements, continuous six-degree-of-freedom ground truth, calibration and frame transformation metadata, and runway geometry and runway-relative states. These outputs supply reference information for characterizing navigation errors and investigating integrity monitoring. Experiments with open-source visual--inertial algorithms support the dataset’s functional usability, while the proposed Sim-to-Real Scene Trustworthiness (SRST) measure provides exploratory evidence of category-relative correspondence between unpaired real and simulated approach images. Together, these results support \textbf{PASS} as a reproducible resource for runway approach navigation and integrity research. 

The current release primarily represents prescribed straight-in approaches. Future work will extend \textbf{PASS} to path and flight state deviations, touchdown, and ground roll phases, while retaining its calibration and coordinate frame consistency.

\section*{Data availability}

The \textbf{PASS} dataset and supporting materials will be made publicly available upon publication at \url{https://github.com/zhulang195/PASS_dataset}.

\bibliography{sn-bibliography}% common bib file

\end{document}